\documentclass[lettersize,journal]{IEEEtran}
\usepackage{amsmath,amsfonts}
\usepackage{algorithmic}
\usepackage{algorithm}
\usepackage{array}
\usepackage{textcomp}
\usepackage{stfloats}
\usepackage{url}
\usepackage{verbatim}
\usepackage{graphicx}
\usepackage{cite}

\usepackage[pagebackref,breaklinks,colorlinks]{hyperref}

\usepackage{multirow}
\usepackage{booktabs}
\usepackage{arydshln}

\usepackage{caption}
\usepackage{subcaption}
\usepackage[table,dvipsnames]{xcolor}

\begin{document}

\title{VSFormer: Mining Correlations in Flexible View Set for Multi-view 3D Shape Understanding}

\author{Hongyu Sun, Yongcai Wang,~\IEEEmembership{Member,~IEEE,}
Peng Wang, Haoran Deng, Xudong Cai, Deying Li
\thanks{All authors are with the Department of Computer Science, School of Information, Renmin University of China, 
Beijing 100872, China. Corresponding author: Yongcai Wang. 
{\tt\small \{sunhongyu, ycw, peng.wang, xudongcai, denghaoran, deyingli\}@ruc.edu.cn}}%
}

\markboth{Journal of \LaTeX\ Class Files,~Vol.~5, No.~3, March~2024}%
{Shell \MakeLowercase{\textit{et al.}}: A Sample Article Using IEEEtran.cls for IEEE Journals}

\maketitle

\begin{abstract}
View-based methods have demonstrated promising performance in 3D shape understanding. 
However, they tend to make strong assumptions about the relations between views or learn the multi-view correlations indirectly, 
which limits the flexibility of exploring inter-view correlations and the effectiveness of target tasks. 
To overcome the above problems, this paper investigates flexible organization and 
explicit correlation learning for multiple views. 
In particular, 
we propose to incorporate different views of a 3D shape into 
a permutation-invariant set, referred to as \emph{View Set}, which removes rigid relation assumptions  and facilitates adequate information exchange and fusion among views. 
Based on that, we devise a nimble Transformer model, named \emph{VSFormer}, 
to explicitly capture pairwise and higher-order correlations of all elements in the set. 
Meanwhile, we theoretically reveal a natural correspondence between the 
Cartesian product of a view set and the correlation matrix in the attention mechanism, which supports our model design. 
Comprehensive experiments suggest that VSFormer has better flexibility, 
efficient inference efficiency and superior performance. Notably, 
VSFormer reaches state-of-the-art results on various 3d recognition datasets, including ModelNet40, ScanObjectNN and RGBD.
It also establishes new records on the SHREC'17 retrieval benchmark. 
The code and datasets are available at \url{https://github.com/auniquesun/VSFormer}.
\end{abstract}


\section{Introduction}
\IEEEPARstart{W}{ith} the advancement of 3D perception devices (LiDAR, RGBD camera, etc.), 3D assets like
point clouds, volumetric grids, polygon meshes, RGBD images 
become more and more common in daily life and industrial production~\cite{huang06thin,shapenet2015,gao15active,Objaverse,OmniObject3D}. 
3D object recognition and retrieval are basic requirements for understanding the 3D contents and the development of these technologies will benefit downstream applications like VR/AR/MR, 3D printing, autopilot, etc. 

Existing methods for 3D shape analysis can be roughly divided into three categories according to the input representation: 
(1) point-based~\cite{qi17pointnet,qi17pointnet2,wang19dgcnn,thomas19kpconv,wu19pointconv,zhao19pointweb,liu19rscnn,yan20pointasnl,xiang21curvenet,
mohammadi21pointview,zhao21pt,ma22pointmlp,li22rif}, 
(2) voxel-based~\cite{wu15modelnet,Maturana15voxnet,qi16volumetric,zhou18voxelnet,shao20hcnn}, and 
(3) view-based methods~\cite{su15mvcnn,su18mvcnn-new,feng18gvcnn,you18pvnet,xu18erfanet,wang19rcpcnn,feng19hgnn,Esteves19emv,li19angular,
leng19vdn,ha19SeqViews2SeqLabels,
han193D2SeqViews,chen19veram,ma19learning,zhang19imhl,wei20viewgcn,hamdi21mvtn,wei23viewgcn2,gao23hgnn+,hamdi23voint,xu24mvpnet}. 
Among them, view-based methods recognize a 3D object according to its rendered or projected images, termed \emph{multiple~views}. 
Generally, methods in this line~\cite{su18mvcnn-new,wei20viewgcn,chen21mvt,xu21carnet,hamdi21mvtn,gao23hgnn+,wei23viewgcn2,hamdi23voint,xu24mvpnet} outperform 
the point- and voxel-based counterparts\cite{qi16volumetric,yan20pointasnl,xiang21curvenet,zhao21pt,ma22pointmlp}. 
On one hand, view-based methods benefit from massive image datasets and the advances in image recognition over the past decade. 
On the other hand, the multiple views of a 3D shape contain richer visual and semantic signals than the point or voxel form. 
For example, one may not be able to decide whether two 3D shapes 
belong to the same category by observing them from one view, but the answer becomes clear after seeing other views of these shapes. 
The example inspires a critical question, how to effectively exploit multi-view information for a better understanding of 3D shape. 

\begin{figure}
      \centering
      \subfloat[Independent Views]{
            \includegraphics[width=0.49\linewidth]{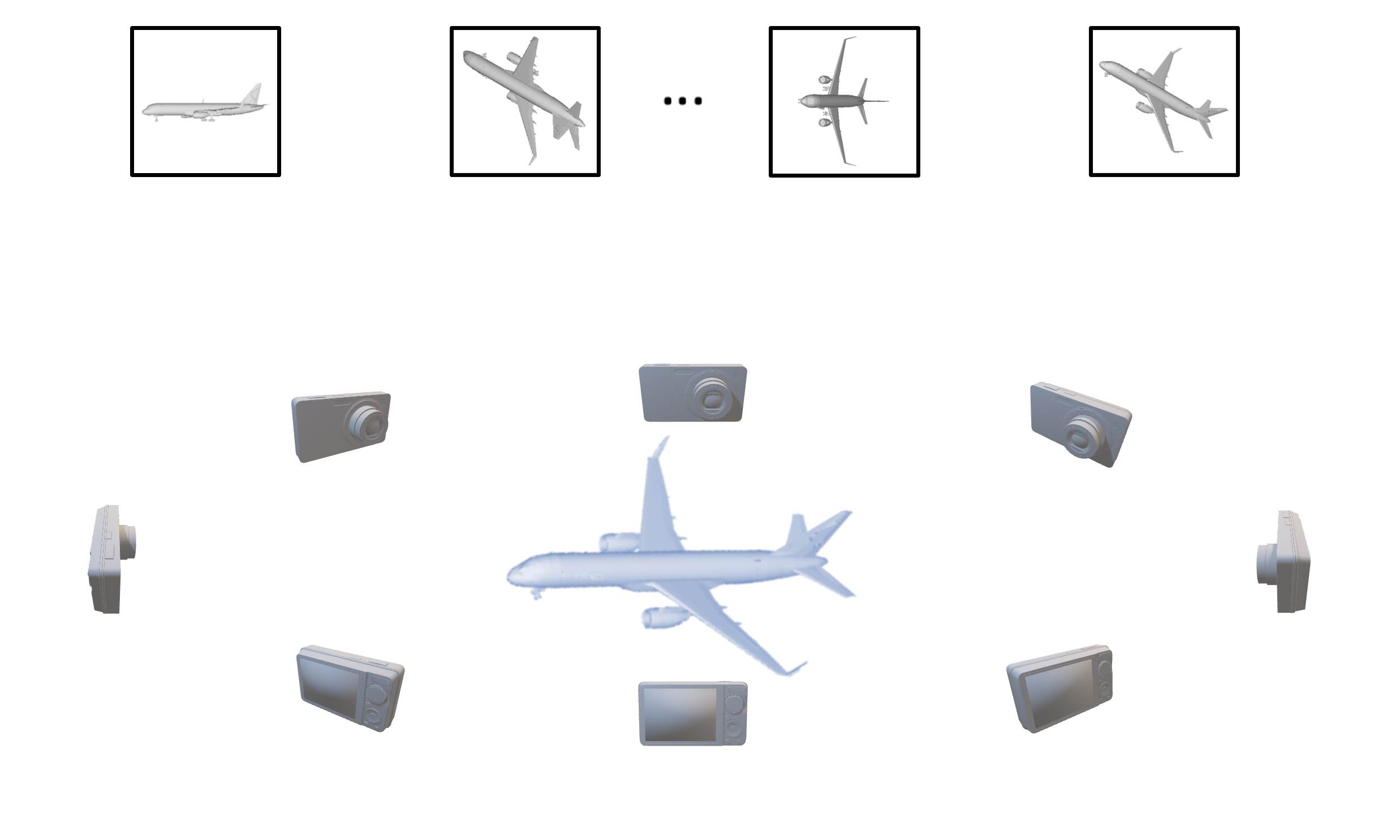}
            \label{fig:independent_view}
      }
      \subfloat[View Sequence]{
            \includegraphics[width=0.49\linewidth]{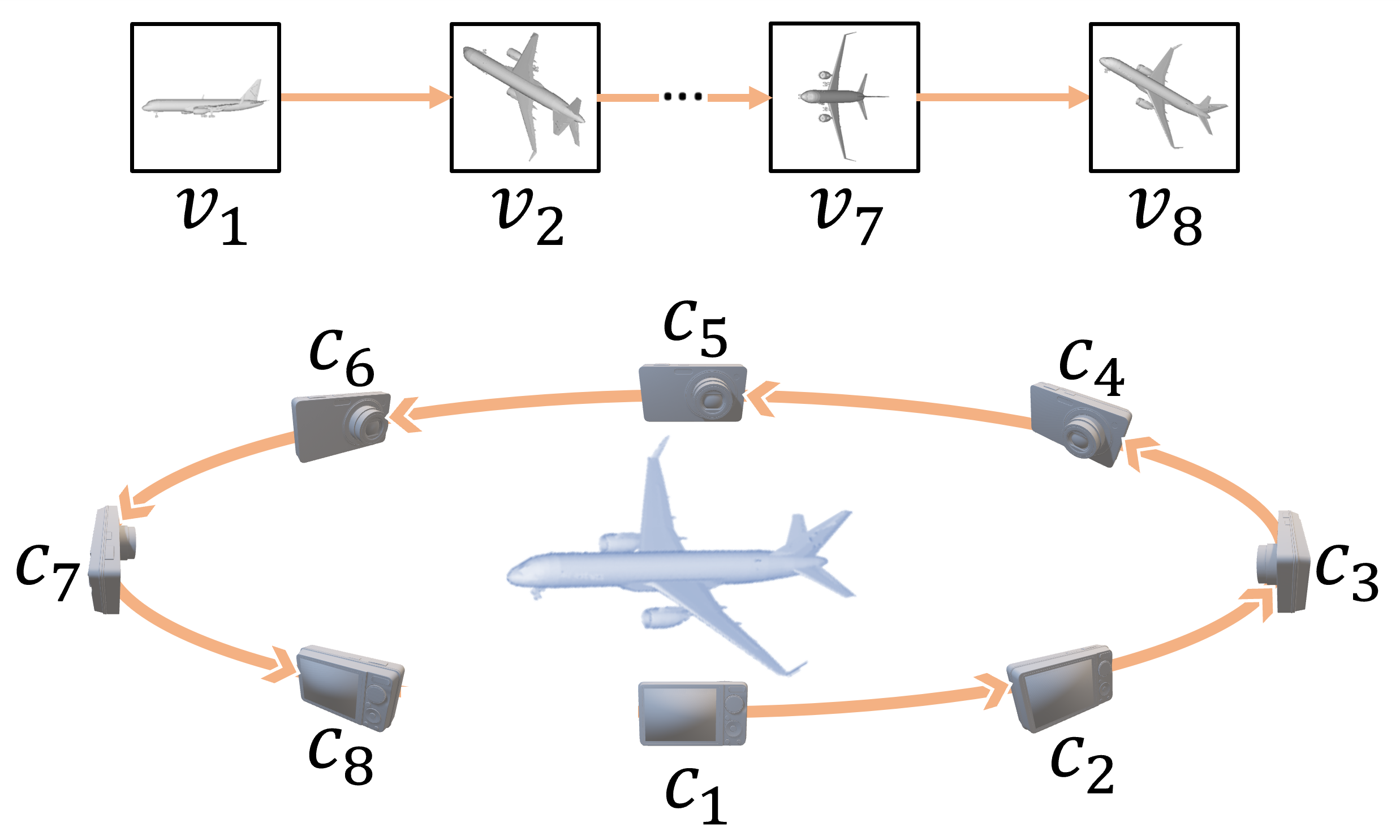}
            \label{fig:view_sequence}
      }\\
      \subfloat[View Graph]{
            \includegraphics[width=0.49\linewidth]{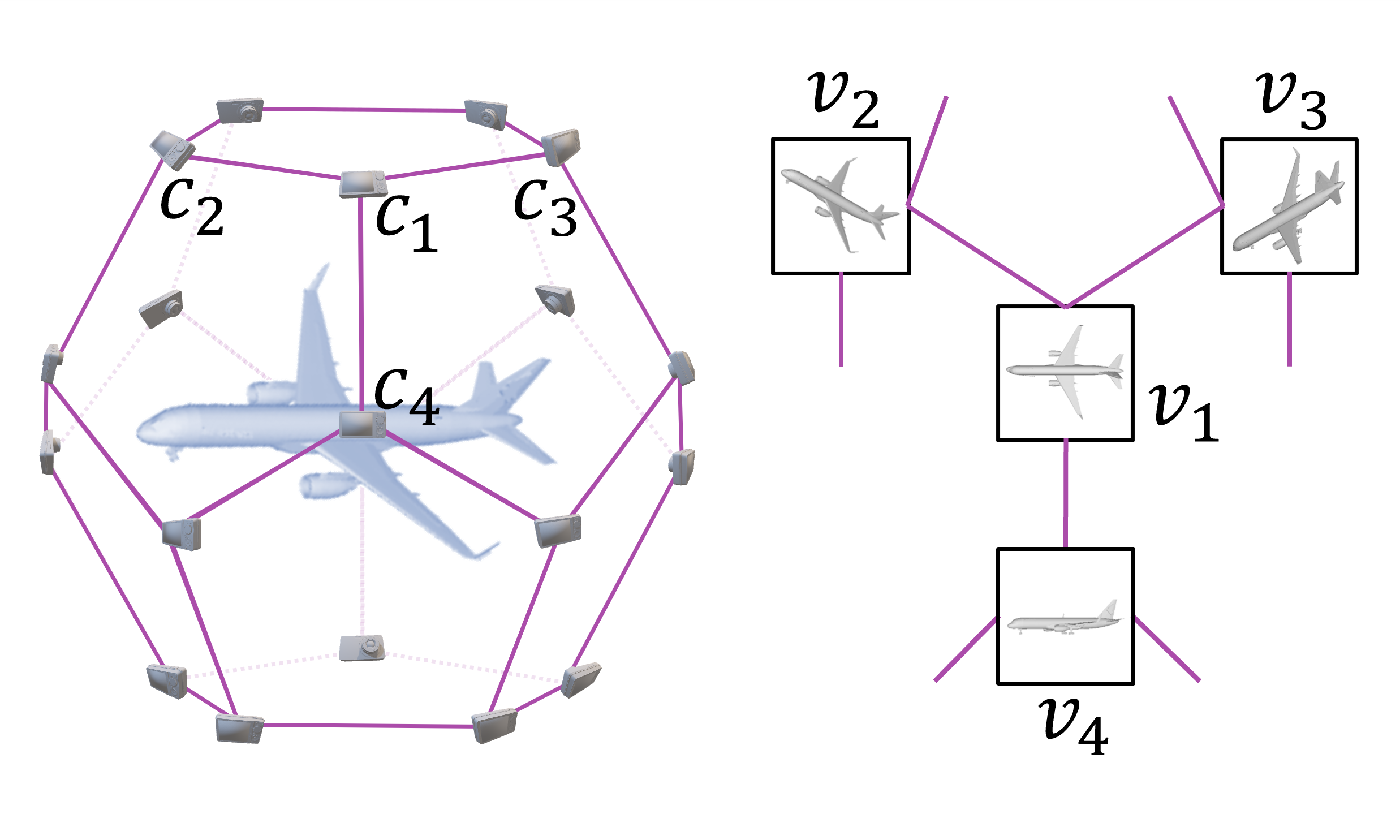}
            \label{fig:view_graph}
      }
      \subfloat[View Set]{
            \includegraphics[width=0.49\linewidth]{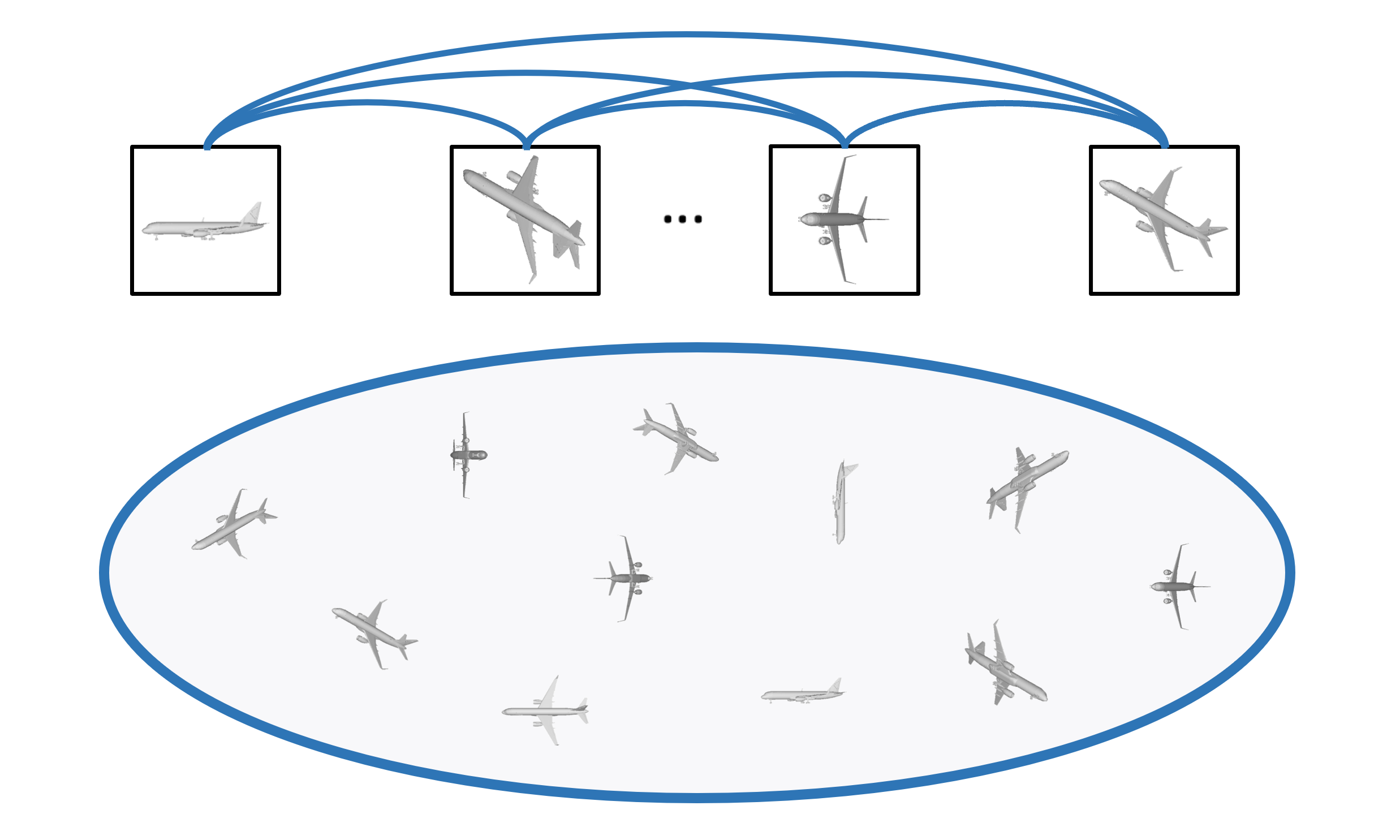}
            \label{fig:view_set}
      }
      \caption{\textbf{A division for multi-view 3D shape analysis methods.} The division is based on how they organize views and 
      aggregate multi-view information. View Set is adopted by VSFormer that the views of a 3D shape are organized in a set.}
      \label{fig:view_structures}
\end{figure}

This paper systematically investigates existing methods on how they aggregate the multi-view information and 
the findings are summarized in Figure~\ref{fig:view_structures}. 
In the early stage, MVCNN~\cite{su15mvcnn} and its follow-up works~\cite{su18mvcnn-new,feng18gvcnn,yu18mhbn,you18pvnet,li19angular,wang19rcpcnn,
leng19vdn,yu21mhbn++} independently process different views of a 3D shape by a shared CNN. 
The extracted features are fused with pooling operation or some variants to form a compact 3D shape descriptor. 
We group these methods into \emph{Independent Views}, shown in Figure~\ref{fig:independent_view}. 
Although the simple design made them stand out at the time, the interaction among different views was insufficient.
In the second category, a growing number of methods model multiple views as a 
sequence~\cite{xu18erfanet, ha19SeqViews2SeqLabels, han193D2SeqViews, chen19veram, ma19learning} to increase information exchange, 
which are grouped into \emph{View Sequence} in Figure~\ref{fig:view_sequence}. 
They deploy RNNs, like GRU~\cite{chung14empirical} and LSTM~\cite{Hochreiter97lstm}, to learn the view relations. 
However, a strong assumption behind \emph{View Sequence} is that the views are collected from a circle around the 3D shape. 
In many cases, the assumption may be invalid since the views can be rendered from random viewpoints, so they are unordered. 
To alleviate this limitation, later methods describe views with a more flexible structure, 
graph~\cite{wei20viewgcn, wei23viewgcn2, xu24mvpnet} or hyper-graph~\cite{zhang19imhl, feng19hgnn, gao23hgnn+}, 
and develop graph convolution networks (GCNs) to propagate features among views, called \emph{View Graph} in Figure~\ref{fig:view_graph}. 
Methods in this category show both flexibility and promising gains, whereas they require to construct 
a view graph for each 3D shape according to the positions of camera viewpoints, which introduces additional computation overheads. 
Meanwhile, the viewpoints may be unknown and the message propagation on the graph may not be straightforward for distant views. 
Some other methods also explore rotations~\cite{Kanezaki18rotationnet,Esteves19emv}, region-to-region relations~\cite{yang19relationnet}, multi-layered height-maps~\cite{sarkar18mlh}, view correspondences~\cite{xu21carnet},
viewpoints selection~\cite{hamdi21mvtn}, voint cloud representations~\cite{hamdi23voint} when recognizing 3D shapes.
They can hardly be divided into the above categories, but multi-view correlations in these 
methods still need to be enriched. 

By revisiting existing works, two aspects are identified critical for improving multi-view 3D shape analysis but are not 
explicitly pointed out in previous literature. 
The first is how to organize the views so they can communicate flexibly and freely. The second is how to 
model multi-view correlations directly and explicitly. It is worth noting that the second ingredient is usually coupled with the first, 
just like GCNs designed for view graphs and RNNs customized for the view sequences. 

In this paper, we propose to organize the multiple views of a 3D shape into a more flexible structure, $e.g.$, \emph{View Set},
shown in Figure~\ref{fig:view_set}, where elements are permutation invariant. This is consistent with the fact that 3D shape understanding is 
actually not dependent on the order of input views. For example, in Figure~\ref{fig:view_sequence}, whether the side view 
is placed first, middle or last in the inputs, the recognition result produced by the model should always be \verb|airplane|. 
Unlike existing methods analyzed above, this perspective removes inappropriate assumptions and restrictions 
about the relations between the views, thus is more practical and reasonable in real-world applications. 

More importantly, a \emph{ViewSet Transformer} (\textbf{VSFormer}) is devised to release the power of multiple views and 
adaptively learn the pairwise and higher-order relations among the views and integrate multi-view information. 
The attention architecture is a natural choice because it aligns with the view set's characteristics. First, 
we theoretically reveal that the Cartesian product of a view set can be formulated by the correlation matrix, which can be 
decomposed into attention operations mathematically. 
Second, the attention mechanism is essentially a set operator and inherently good at capturing correlations between the elements in a set. 
Third, this mechanism is flexible enough that it makes minimal assumptions about the inputs, which matches our expectation
that there are no predefined relations or restrictions for views. 
Overall, the proposed approach presents a one-stop solution that directly captures the correlations of all view pairs in the set, 
which promotes the flexible and free exchange of multi-view information.

Several critical designs are presented in VSFormer. 
(1) The position encodings of input views are removed since views are permutation invariant. 
(2) The class token is removed because it is irrelevant to capturing the correlations of view pairs in the set. 
(3) The number of attention blocks is greatly reduced as the size of a view set is relatively small ($\leq$ 20 in most cases).  

The details of the proposed approach will be elaborated in Section~\ref{sec:method}. 
Systematic experiments suggest that VSFormer around the flexible set and explicit relation grasping 
unleashes astonishing capabilities and obtains new records in downstream tasks. 
In short, the contributions of this paper include: 

\begin{itemize}
   \item We identify two key aspects of multi-view 3D shape understanding, 
   organizing views reasonably and modeling their relations explicitly,  
   albeit they are critical for performance improvement but absent in previous literature. %
   \item We propose 
   a Transformer-based model, named VSFormer, to capture the correlations of all view pairs directly for better 
   multi-view information exchange and fusion. At the same time, a theoretical analysis 
   is accompanied to support such a design. 
   \item Extensive experiments demonstrate the superb performances of the proposed approach and the ablation studies 
   shed light on the various sources of performance gains. 
   For the recognition task, VSFormer reaches 98.4\%(+4.1\%), 95.9\%(+1.9\%), 98.8\%(+1.1\%) overall accuracy 
   on RGBD, ScanObjectNN, ModelNet40, respectively. The results surpass all existing methods and achieve new state of the arts. 
   For 3D shape retrieval, VSFormer also sets new records in multiple dimensions on the SHREC'17 benchmark.
\end{itemize} 

\begin{figure*}[t]
	\begin{center}
		\includegraphics[width=\linewidth]{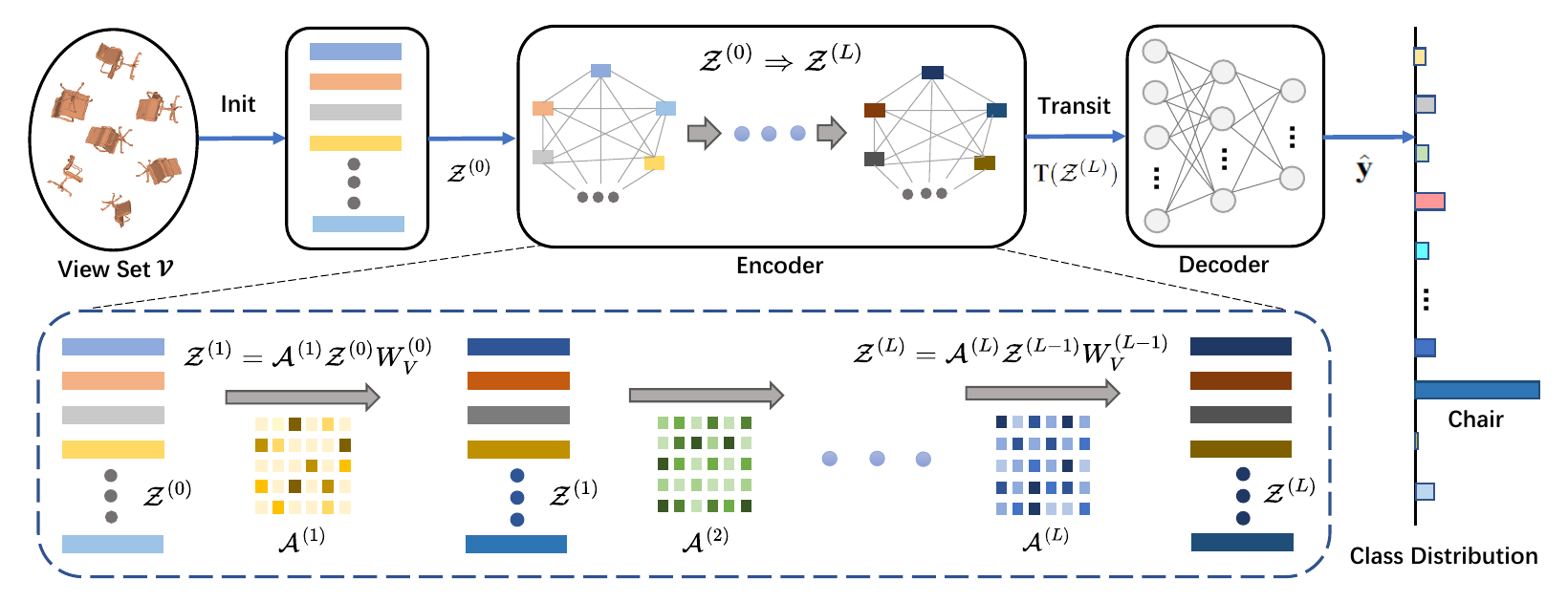}
	\end{center}
	\caption{\textbf{The overall architecture of VSFormer.} It consists of 4 modules: Initializer (Init), Encoder, Transition (Transit) and Decoder.
   Encoder is responsible for grasping pairwise and higher-order correlations of views in a set.}
	\label{fig:architecture}
\end{figure*}

\section{Related Work}
In this section, we review the multi-view 3D shape analysis methods, explore the deployment of set and attention in these methods, 
and discuss the latest progress in the field. 

\subsection{Multi-view 3D Shape Analysis} 
Existing methods aggregate multi-view information for 3D shape understanding in different ways. 

\subsubsection{Independent Views} Early work like MVCNN series~\cite{su15mvcnn,su18mvcnn-new} and its follow-up works~\cite{feng18gvcnn,yu18mhbn,
you18pvnet,li19angular,wang19rcpcnn,leng19vdn,yu21mhbn++}
extract view features independently using a shared CNN, then fuse the extracted features using the pooling operation or some variants. 
The simple strategy may discard a lot of useful information and the views are not well treated as a whole 
thus information flow among views needs to be increased. 

\subsubsection{View Sequence} Researchers perceive the problems and propose various descriptions to incorporate multiple views 
of a 3D shape into specific data structures. For example,
RNN-based~\cite{xu18erfanet, ha19SeqViews2SeqLabels, han193D2SeqViews, chen19veram, ma19learning} and ViT-based~\cite{chen21mvt, lin23mrvanet} methods 
are proposed to operate on the view sequence. 

\subsubsection{View Graph} The graph-based models~\cite{feng19hgnn, zhang19imhl, wei20viewgcn, wei23viewgcn2, gao23hgnn+, xu24mvpnet} assume 
the relations among views as graphs and develop GCNs to capture multi-view interaction. 
However, message propagation between distant nodes on a view graph may not be straightforward and graph construction leads to 
additional computation overheads. 

\subsubsection{View Set} This paper presents a more flexible and practical structure, \emph{View Set}, which neither makes
assumptions about views nor introduces additional overheads. 
Based on that, a view set attention model is devised to adaptively grasp the correlations for all view pairs. 

Some other methods also explore rotations~\cite{Kanezaki18rotationnet,Esteves19emv}, region-to-region relations~\cite{yang19relationnet}, multi-layered height-maps representations~\cite{sarkar18mlh}, view correspondences~\cite{xu21carnet},
viewpoints selection~\cite{hamdi21mvtn}, voint cloud representations~\cite{hamdi23voint} when analyzing 3D shapes. Their multi-view interaction still needs to be strengthened. 

\subsection{Set in Multi-view 3D Shape Analysis} 
Previous works also mention ``set" in multi-view 3D shape analysis. 
But they basically refer to different concepts from the proposed one. 
For instance, RCPCNN~\cite{wang19rcpcnn} introduces a dominant set clustering and pooling
module to improve MVCNN~\cite{su15mvcnn}.
Johns \textit{et al.}~\cite{johns16pairwise} decompose a sequence of views into a set of view pairs. They classify each pair independently and weigh the 
contribution of each pair. MHBN~\cite{yu18mhbn} considers patches-to-patches (set-to-set) similarity of different views
and aggregates local features using bilinear pooling. Yu \textit{et al.} extend MHBN by introducing VLAD layer~\cite{yu21mhbn++},  
where the similarity between two sets of local patches is calculated by exploiting bilinear and VLAD pooling operations, 
while our view set perspective provides a foundation for learning the correlations of all view pairs adaptively. 

\subsection{Attention in Multi-view 3D Shape Analysis} 
The attention mechanisms have been embedded in existing multi-view 3D shape analysis methods but vary in 
motivation, practice and effectiveness. VERAM~\cite{chen19veram} uses a recurrent attention model to select
a sequence of views to classify 3D shapes. SeqViews2SeqLabels~\cite{ha19SeqViews2SeqLabels} introduces the attention 
mechanism to increase the discriminative ability for the RNN-based model and reduces the effect of selecting the first view position.  
3D2SeqViews~\cite{han193D2SeqViews} proposes hierarchical attention to incorporate view-level and class-level importance
for 3D shape analysis. Nevertheless, there are three points worth noting for the attention in the above methods. 
Firstly, the attention modules in these methods have nothing to do with \emph{view set perspective} and 
are not designed for handling an unordered view set.
Secondly, these modules differ from the multi-head self-attention in standard Transformer~\cite{vaswani17transformer}. 
Thirdly, previous methods equipped with attention modules do not seem to produce satisfactory performances. 

Another work MVT~\cite{chen21mvt} also explores the attention architecture for view-based 3D recognition. 
However, MVT is inspired by the success of ViT~\cite{dosovitskiy21vit} and simply applies ViT to the views without modification. 
The position encodings in MVT are preserved. Thus the method is also irrelevant to the idea of organizing views in an unordered set. 
Besides, MVT deploys a ViT to extract patch-level features and adopts another ViT to learn the correlations of all patches in different views. 
In contrast, VSFormer shows it is unnecessary to take the patch-level interactions into account to achieve better results, 
thus the computation budgets are significantly reduced. 
Recent work MRVA-Net~\cite{lin23mrvanet} investigates multi-range view aggregation with ViT-based feature fusion for 3D shape retrieval. 
This method belongs to the category of \emph{View Sequence} as it assumes the views are rendered along a circle. 
The short, mid and long ranges are defined by human priors and the model may be sensitive to the range definition. 
Then dilated convolutions are conducted on the CNN-initialized view features 
to attain multi-range features. After that, MRVA-Net applies ViT without adaptation to fuse the multi-range features.
Instead, our approach is more flexible and efficient by directly processing an unordered set of views in parallel, 
without the sequence assumption, multiple convolution operations, and hard-encoded ranges for views. 

\subsection{Latest Progress in Multi-view 3D Shape Analysis}
Here we discuss several very recent works and highlight the differences with our method. 
Hamdi $et~al.$ propose a novel Voint Cloud representation and 
develop VointNet~\cite{hamdi23voint} to conduct multiple 3D tasks. 
This model is built on conventional 2D backbones and combines multi-view information with 3D point clouds 
while our model is devised upon standard attention mechanism and only exploits views as inputs. 
HGNN$^{+}$~\cite{gao23hgnn+} proposes hyperedge groups and hypergraph convolution to explore multi-modal data correlation.
In our cases, we only have views and do not emphasize multi-modal data correlation. 
View-GCN++~\cite{wei23viewgcn2} wants to deal with rotation sensitivity and upgrades the prior version by 
developing local attentional graph convolution and rotation robust view sampling. 
Instead, our method directly processes aligned or rotated views by exploring the correlations of all view pairs in parallel,
which demonstrates superior performances. 
MVPNet~\cite{xu24mvpnet} improves View-GCN~\cite{wei20viewgcn} by generating an ordered path on the view graph 
and aggregating the ordered features along the path with ViT~\cite{dosovitskiy21vit}. 
It falls into the View Graph category and may suffer from the shortcomings as we analyzed above,
$e.g.$, indirect relation modeling and additional overheads introduced by graph construction and 
path generation for each 3D object. 

\section{Methodology}
\label{sec:method}

In this section, we firstly formulate the problem of multi-view 3D shape analysis 
based on the view set, then elaborate on the devised VSFormer and how it handles a set of views. 

\subsection{Problem Formulation}

\subsubsection{View Set} The views of a 3D shape refer to its rendered or projected RGB images. For example, 
a 3D shape $\mathcal{S}$ corresponds to views $v_1, v_2, \dots, v_M \in \mathbb{R}^{H\times W\times C}$, 
where $M$ is the number of views and $H\times W\times C$ indicates the image size. 
In \emph{our perspective}, the views of $\mathcal{S}$ form a set $\mathcal{V} = \{v_1, v_2, \dots, v_M\}$, 
where elements are permutation invariant. 
Thus, $\mathcal{V}_{\pi} = \{v_{\pi_{(1)}}, v_{\pi_{(2)}}, \dots, v_{\pi_{(M)}}\}$ is always equivalent to 
$\mathcal{V}$ when $\pi_{(1)}, \pi_{(2)}, \dots, \pi_{(M)}$ is a random permutation of 1, 2, $\dots$, $M$. 

In fact, organizing different views in order (view sequence) is a special case of random permutation.
Random permutation does not introduce additional overheads compared to view sequence, 
and considerably saves computation budgets compared to graph construction in view graph methods. 

\subsubsection{3D Shape Recognition \& Retrieval}
In many cases, 3D shape retrieval can be regarded as a classification problem~\cite{savva17shrec17}. 
It aims to find the most relevant shapes to the query. Meanwhile, the relevance is defined according to 
the query's ground truth class and subclass, which means if a retrieved shape has the same class and 
subclass as the query, they match perfectly. Therefore, the tasks of 3D shape retrieval and recognition can be unified by 
predicting a category distribution $\hat{\textbf{y}} \in \mathbb{R}^{K}$ of the target shape $\mathcal{S}$, 
where $K$ is the number of 3D shape categories. 
In this paper, we design a view set attention model $\mathcal{F}$ to %
predict the distribution. 
The input of $\mathcal{F}$ is a view set $\mathcal{V} \in \mathbb{R}^{M\times H\times W\times C}$ of the shape $\mathcal{S}$, 
and the output is a class distribution $\hat{\textbf{y}} = \mathcal{F}(\mathcal{V})$. 

\subsection{View Set Attention Model}
The proposed model aims to facilitate flexible communication and adequate information fusion among views in a set. 
The overall architecture of VSFormer is presented in Figure~\ref{fig:architecture}.

The input of VSFormer is a view set $\mathcal{V} = \{v_1, \dots, v_M\}$. 
First, we initialize $\mathcal{V}$ with lightweight modules ($e.g.$, ResNet18~\cite{he16resnet}, AlexNet~\cite{krizhevsky12alexnet}) 
to map the views to 
hidden representations $\mathcal{Z}^{(0)} = \{\textbf{z}_1^{(0)}, \dots, \textbf{z}_M^{(0)}\} \in \mathbb{R}^{M \times D}$, 
where $\textbf{z}_i^{(0)}$ 
is a $D$-dimensional encoding of the view $v_i$. 
So $\mathcal{Z}^{(0)}$ contains information from $M$ independent views, without any clue of their correlations. 
Second, to enrich the interaction and fusion of multi-view information, the designed model computes the correlations 
of all view pairs directly through iterative attention. 

We show that 
(1) in theory, there is a natural correspondence between the view pairs
in a set and the correlation matrix in a standard attention model. 
(2) in practice, the proposed model can be constructed according to the view set and attention theory.

\subsubsection{View Set and Attention Theory}
\label{subsubsec:view_set_att_theory}
The first problem is how to represent view pairs in a set. 
For the initialized $\mathcal{Z}^{(0)}$, all view pairs can be formulated by its \emph{Cartesian product}
$\mathcal{P}^{(0)} = \mathcal{Z}^{(0)}\times \mathcal{Z}^{(0)} = \{\ (\textbf{z}_i^{(0)}, \textbf{z}_j^{(0)})\ |\ i,j \in 1,\dots,M\}$. 
Let us denote $p_{i,j}^{(0)} = (\textbf{z}_i^{(0)}, \textbf{z}_j^{(0)})$, so all view pairs in $\mathcal{Z}^{(0)}$ can be expressed 
with $\mathcal{P}^{(0)} = \{\ p_{i,j}^{(0)}\ |\ i,j \in 1,\dots,M\}$. 

\newtheorem{theorem}{Theorem}
\begin{theorem}[Correspondence between the Cartesian product of a view set and the correlation matrix in the attention mechanism]
The Cartesian product $\mathcal{P}$ of a view set $\mathcal{V}$ can be formulated by a correlation 
matrix $\mathcal{A}$ and computed by the attention mechanism. %
\label{theorem:theo1}
\end{theorem}

\emph{Proof} is provided in the subsection A of the Supplementary Material. We further elaborate on the theorem from the following three aspects. 

\noindent\textbf{(i) Pairwise Correlations in a View Set.}
Generally, the attention model characterizes the pairwise correlations of different elements by a correlation matrix. 
The model receives an input $\mathcal{I}$ that consist of $N$ elements, 
$\textbf{e}_1, \dots, \textbf{e}_N$, where $\textbf{e}_i$ is a $E$-dimensional vector. 
$\mathcal{I}$ is regarded as a set $\{\textbf{e}_1, \dots, \textbf{e}_N\}$ as the model is unaware of the order of the elements. 
Thereby, the pairwise correlations for $\mathcal{I}$ learned by the attention mechanism can be represented 
by a correlation matrix $\mathcal{A}^{(1)} = \{\ a_{i,j}^{(1)}\ |\ i,j \in 1,\dots,N\}$, 
shown in Eq.~\ref{eq:attention_matrix}, 
where $a_{i,j}^{(1)}$ is the attention score that $\textbf{e}_i$ receives from $\textbf{e}_j$. 
\begin{equation}
   \mathcal{A}^{(1)} = 
   \begin{bmatrix}
      a_{1,1}^{(1)} & a_{1,2}^{(1)} & \dots & a_{1,N}^{(1)} \\
      a_{2,1}^{(1)} & a_{2,2}^{(1)} & \dots & a_{2,N}^{(1)} \\
      \dots & \dots & \dots & \dots \\
      a_{N,1}^{(1)} & a_{N,2}^{(1)} & \dots & a_{N,N}^{(1)} 
   \end{bmatrix}
   \label{eq:attention_matrix}
\end{equation}

We define $\mathcal{A}^{(1)}$ as the \emph{first-order correlation} matrix. 
It is noticed that $\mathcal{A}^{(1)}$ has same form as 
the \emph{Cartesian product} $\mathcal{P}^{(0)}$. 
Let $\textbf{a}_{i}^{(1)} = (a_{i,1}^{(1)}, \dots, a_{i,N}^{(1)})$ and $i \in 1,\dots,N$, $\textbf{a}_{i}^{(1)}$ represents 
the correlations that the $i$th element receives from all elements in $\mathcal{I}$. 
Hence, $\mathcal{A}^{(1)}$ can be further converted into another form in Eq.~\ref{eq:attention_matrix_new} 
and $\textbf{a}_{i}$ can be calculated with Eq.~\ref{eq:attention_matrix_row}. 
\begin{gather}
   \begin{split}
      \mathcal{A}^{(1)} &= 
      \begin{bmatrix}
         \textbf{a}_{1}^{(1)} & \textbf{a}_{2}^{(1)} & \dots & \textbf{a}_{N}^{(1)}
      \end{bmatrix}^{\textrm{T}}\\
   \end{split}
   \label{eq:attention_matrix_new}
\end{gather}
\begin{equation}
   \begin{split}
      &\textbf{a}_{i}^{(1)} = \textrm{Norm}(Q_i^{(0)} {K^{(0)}}^\textrm{T} /\ {\tau})\\
      Q_i^{(0)} &= \textbf{e}_i W_Q^{(0)}\ \quad K^{(0)} = \mathcal{I} W_K^{(0)}
   \end{split}
   \label{eq:attention_matrix_row}
\end{equation}
Here $\tau$ is a temperature coefficient to adapt the product and Norm is a normalized function to ensure the attention 
scores are in the range of [0,1]. 
Both $W_Q^{(0)}$ and $W_K^{(0)} \in \mathbb{R}^{E\times E}$ are learnable embeddings to project the input $\mathcal{I}$. 

Due to $\mathcal{P}^{(0)}$ and $\mathcal{A}^{(1)}$ having the same mathematical form,
it is easy to transfer the above process to capture the pairwise correlations of all views in a set. 
The only thing we need to do is make $N = M$, $E = D$ and $\mathcal{I} = \mathcal{Z}^{(0)}$.

\noindent\textbf{(ii) Injecting the Correlations into Views.} 
Once the \emph{first-order correlations} $\mathcal{A}^{(1)}$ are obtained, we can inject this kind of knowledge into the initialized 
$\mathcal{Z}^{(0)}$ using Eq.~\ref{eq:first_order_view_rep}, to enable information flow between views. 
The idea is to update each element of $\mathcal{Z}^{(0)}$ according to its correlations with other elements, resulting in 
$\mathcal{Z}^{(1)} = \{\textbf{z}_1^{(1)}, \dots, \textbf{z}_M^{(1)}\} \in \mathbb{R}^{M\times D}$. 
\begin{equation}
   \mathcal{Z}^{(1)} = \mathcal{A}^{(1)}\mathcal{Z}^{(0)}W_V^{(0)}
   \label{eq:first_order_view_rep}
\end{equation}
Here $W_V^{(0)} \in \mathbb{R}^{D\times D}$ is a learnable embedding that is to map the initialized $\mathcal{Z}^{(0)}$.
The new representations $\mathcal{Z}^{(1)}$ achieve a basic understanding of the relations
among views thus take the first step toward multi-view information fusion. 
We call $\mathcal{Z}^{(1)}$ the \emph{first-order representations} of the view set $\mathcal{V}$. 

However, the first-order representations may not be sufficient to grasp various view relations in 
complex scenarios. It is expected that the \emph{higher-order} interactions among views are also adaptively explored. 

\noindent\textbf{(iii) Higher-order Correlations in a View Set.} 
The attention model can go beyond capturing pairwise correlations for elements in a set. 
Assuming there are a total of $L$ attention blocks in the model, 
since the correlation matrix $\mathcal{A}^{(\ell)}$ in $\ell$th block is always constructed based on $\mathcal{A}^{(\ell-1)}$, 
the higher-order interaction can be learned by deepening the attention blocks. 
We derive the $\ell$th-order multi-view representations $\mathcal{Z}^{(\ell)}$ using Eq.~\ref{eq:n_order_view_rep}, 
where $W_Q^{(\ell-1)}, W_K^{(\ell-1)}$ and $W_V^{(\ell-1)}$ are learnable embeddings of shape $\mathbb{R}^{D\times D}$ 
to transform $\mathcal{Z}^{(\ell-1)}$, $\ell \in 1, \dots, L$. 
\begin{gather}
   \centering
   \begin{split}
      \mathcal{Z}^{(\ell)} &= \mathcal{A}^{(\ell)}\mathcal{Z}^{(\ell-1)}W_V^{(\ell-1)}\\
      \mathcal{A}^{(\ell)} &= \textrm{Norm}(Q^{(\ell-1)}{K^{(\ell-1)}}^\textrm{T} /\ \tau) \\
      Q^{(\ell-1)} &= \mathcal{Z}^{(\ell-1)}W_Q^{(\ell-1)}\\
      K^{(\ell-1)} &= \mathcal{Z}^{(\ell-1)}W_K^{(\ell-1)}\\
   \end{split}
   \label{eq:n_order_view_rep}
\end{gather}
By going through $L$ attention blocks in the view set encoder, the representations $\mathcal{Z}^{(\ell)}$ 
iteratively compute the correlation matrix and update themselves with the latest knowledge to obtain 
higher-order understanding of the correlations among views.

\subsubsection{Constructing VSFormer}
According to the above analysis, VSFormer can be built with the following modules: Initializer, Encoder, Transition and Decoder, 
shown in Figure~\ref{fig:architecture}. 

Lightweight neural networks can serve as the initializer ($e.g.$, AlexNet~\cite{krizhevsky12alexnet}). 
The encoder receives the initialized view set $\mathcal{Z}^{(0)} \in \mathbb{R}^{M\times D}$ and processes it with $L$ attention blocks. 
Each attention block stacks the multi-head self-attention\cite{vaswani17transformer} (MSA) and MLP layers with residual connections. 
LayerNorm (LN) is deployed before MSA and MLP, whereas Dropout (DP) is applied afterward. 
The procedure in the $\ell$th block is summarized by Eq.~\ref{eq:attn_block}, where $\ell = 1, \dots, L$. 
\begin{equation}
   \centering
   \begin{split}
      \hat{\mathcal{Z}}^{(\ell)} &= \textrm{DP}(\textrm{MSA}(\textrm{LN}(\mathcal{Z}^{(\ell-1)}))) + \mathcal{Z}^{(\ell-1)} \\
      \mathcal{Z}^{(\ell)} &= \textrm{DP}(\textrm{MLP}(\textrm{LN}(\hat{\mathcal{Z}}^{(\ell)}))) + \hat{\mathcal{Z}}^{(\ell)}
   \end{split}
	\label{eq:attn_block}
\end{equation}

Note the input of each attention block is not equipped with the \emph{position encoding} as in standard Transformer~\cite{vaswani17transformer}, 
since the views are permutation invariant in the set. Also, VSFormer does not insert the \emph{class token} in the input as the goal is 
to grasp the correlations within views rather than learning the relations between the class token and views. 
Surprisingly, a lightweight view set encoder (2.7M parameters) that is only composite of 
two attention blocks can work quite well (99.0\% overall accuracy), validated by extensive experiments in Section~\ref{sec:experiments}. 

For 3D shape recognition or retrieval tasks, it is necessary to convert the learned higher-order view set representations 
into an expressive descriptor $\textbf{d} \in \mathbb{R}^{G}$ via a transition module ($e.g.$, concatenation of max and mean pooling). 
Then the descriptor is processed by a decoder to generate the prediction $\hat{\textbf{y}} \in \mathbb{R}^{K}$. 
\begin{align}
   \textbf{d} &= \textrm{Transition}(\mathcal{Z}^{(L)}) \\
   \hat{\textbf{y}} &= \textrm{Decoder}(\textbf{d})
   \label{eq:transsition_decoder_loss}
\end{align}
The objective to be optimized is defined as \emph{Cross Entropy} loss for 3D shape recognition, where $\textbf{y}_i$  
is the predicted class distribution of $i$th object and $\theta_{W}$ denotes all learnable parameters in the model. 
\begin{equation}
   \mathcal{L}_{CE}(\textbf{y}_i, \hat{\textbf{y}}_i; \theta_{W}) = \sum\nolimits_{i} -\textbf{y}_i\log\hat{\textbf{y}}_i
\end{equation}

These modules have various design choices, 
we will examine the design choices of each component through the ablation studies in Section~\ref{subsec:ablation_studies}. 

\subsection{Implementation Details}
\noindent\textbf{Architecture.}
For Initializer, we adopt lightweight CNNs. There are several candidates (AlexNet, ResNet18, \textit{etc.}) and we will compare them later. 
$\mathcal{V}$ is instantiated as a random permutation of the views, 
$v_i \in \mathcal{V}$ is mapped to a $D$=512 dimensional vector through Initializer. 
For Encoder, there are $L$=4 attention blocks and within each block, the MSA layer has 8 attention heads and the widening
factor of the MLP hidden layer is 2. 
The normalized function Norm is defined as softmax($\cdot$) and the temperature coefficient $\tau$ is set to $\sqrt{\frac{D}{8}}$. 
The Transition module converts $\mathcal{Z}^{(L)}$ into a $G$=1024 dimensional descriptor $\textbf{d}$. 
Finally, the descriptor is projected to a category distribution by Decoder, which is a 2-layer MLP of 
shape \{1024, 512, $K$\}. The design choices are verified by ablated studies in Section~\ref{subsec:ablation_studies}. 

\noindent\textbf{Optimization Strategy.}
The optimization objective $\mathcal{L}_{CE}$ has a label smooth of 0.1. 
Following previous methods~\cite{su18mvcnn-new,wei20viewgcn}, 
the learning is divided into two stages. In the first stage, the initializer is individually trained on the target dataset
for 3D shape recognition. The purpose is to provide good initializations for views.  
In the second stage, the pre-trained initializer is loaded and jointly optimized with other modules on the same dataset. 
Experiments in Figure~\ref{fig:VSFormer_training_strategy} show this strategy will significantly improve performance in a shorter period. 

\begin{table}[t]\small
   \begin{center}
      \caption{Comparison of 3D shape recognition on ModelNet40.} 
      \begin{tabular}{l l c c}
      \toprule
      \multirow{2}{*}{Method} & \multirow{2}{*}{Input} & Class Acc. & Inst. Acc.\\
      & & (\%) & (\%)\\
      \midrule
      3DShapeNets~\cite{wu15modelnet} & \multirow{5}{*}{Voxels} & 77.3 & -- \\
      VoxNet~\cite{Maturana15voxnet} &  & 83.0 & -- \\
      VRN Ensemble~\cite{brock16vrn} &  & -- & 95.5 \\
      HCNN~\cite{shao20hcnn} & & -- & 89.4 \\
      MVCNN-MR~\cite{qi16volumetric} &  & 91.4 & 93.8 \\\hdashline
      PointNet++~\cite{qi17pointnet2} & \multirow{8}{*}{Points} & -- & 91.9 \\
      DGCNN~\cite{wang19dgcnn} &  & 90.2 & 92.9 \\
      RSCNN~\cite{liu19rscnn} &  & -- & 93.6 \\
      KPConv~\cite{thomas19kpconv} &  & -- & 92.9 \\
      SimpleView~\cite{goyal21revisiting} &  & 90.5 & 93.0 \\
      CurveNet~\cite{xiang21curvenet} &  & -- & 93.8 \\
      PointMLP~\cite{ma22pointmlp} &  & 91.3 & 94.1 \\
      RIF~\cite{li22rif} & & -- & 89.3 \\\hdashline
      MVCNN~\cite{su15mvcnn} & \multirow{22}{*}{Views} & 90.1 & 90.1 \\
      MVCNN-new~\cite{su18mvcnn-new} &  & 92.4 & 95.0 \\
      MHBN~\cite{yu18mhbn} &  & 93.1 & 94.7 \\
      GVCNN~\cite{feng18gvcnn} &  & 90.7 & 93.1 \\
      RCPCNN~\cite{wang19rcpcnn} &  & -- & 93.8 \\
      RN~\cite{yang19relationnet} & & 92.3 & 94.3 \\
      3D2SeqViews~\cite{han193D2SeqViews} &  & 91.5 & 93.4 \\
      SV2SL~\cite{ha19SeqViews2SeqLabels} &  & 91.1 & 93.3 \\
      VERAM~\cite{chen19veram} &  & 92.1 & 93.7 \\
      Ma \textit{et al.}~\cite{ma19learning} &  & -- & 91.5 \\
      iMHL~\cite{zhang19imhl} &  & -- & 97.2 \\
      HGNN~\cite{feng19hgnn} &  & -- & 96.7 \\
      HGNN$^+$~\cite{gao23hgnn+} &  & -- & 96.9 \\
      View-GCN~\cite{wei20viewgcn} &  & {96.5} & 97.6 \\
      View-GCN++~\cite{wei23viewgcn2} &  & {96.5} & 97.6 \\
      DeepCCFV~\cite{huang19deepccfv} &  & -- & 92.5 \\
      EMV~\cite{Esteves19emv} &  & 92.6 & 94.7 \\
      RotationNet~\cite{Kanezaki18rotationnet} &  & -- & 97.4 \\
      MVT~\cite{chen21mvt} &  & -- & 97.5 \\
      CARNet~\cite{xu21carnet} &  & -- & {97.7} \\
      MVTN~\cite{hamdi21mvtn} &  & 92.2 & 93.5 \\
      MVPNet~\cite{xu24mvpnet} & & 96.8 & 97.9 \\
      \midrule
      \textbf{VSFormer} & Views & \textbf{98.9} & \textbf{98.8}\\
      \bottomrule
      \end{tabular}
      \label{tab:recog_modelnet40}
   \end{center}
\end{table}

\noindent\textbf{Network Training.}
For Initializer, we train it 30 epochs on the target dataset using SGD~\cite{ruder2016overview}, with an initial learning rate of 0.01 and 
CosineAnnealingLR scheduler. After that, the pre-trained weights of Initializer are loaded into VSFormer to be optimized with other modules jointly. 
Specifically, VSFormer is trained 300 epochs on the target dataset using AdamW~\cite{loshchilov2018decoupled}, 
with an initial peak learning rate of 0.001 and CosAnnealingWarmupRestartsLR scheduler~\cite{Katsura21cawwr}. The restart interval is 100 epochs 
and the warmup happens in the first 5 epochs of each interval. 
The learning rate increases to the peak linearly during warmup and the peak decays by 40\% after each interval. 
The learning rate curve is visualized in Figure~\ref{fig:VSFormer_lr}. 

\section{Experiments}
\label{sec:experiments}
In this section, VSFormer is evaluated on 3D shape recognition and retrieval tasks. 
Then we conduct controlled experiments to examine the design choices of the proposed method. 

\subsection{3D Shape Recognition}
\label{subsec:exp_recognition}

\noindent\textbf{Datasets.} 
We conduct 3D shape recognition on three datasets, ModelNet40~\cite{wu15modelnet}, ScanObjectNN~\cite{uy19sonn} and RGBD~\cite{lai11rgbd}. 

\begin{itemize}
   \item ModelNet40 includes 12,311 objects across 40 categories and 
we use its rendered version as in previous work~\cite{su18mvcnn-new,wei20viewgcn}, 
where each object corresponds to 20 views. 
   \item ScanObjectNN is collected from real-world scans and poses great challenges to existing methods. 
There are 2,902 objects distributed in 15 categories. We follow previous work~\cite{wei20viewgcn,wei21cvr,hamdi21mvtn,wei23viewgcn2} to adopt the OBJ\_ONLY split
and render 20 views for each object. To ease reproduction, we wrote a detailed instruction to explain the rendering procedure, 
and please refer to this blog\footnote{\url{https://auniquesun.github.io/2023-06-16-multi-view-rendering/}}.
   \item RGBD is a large-scale, hierarchical multi-view object dataset~\cite{lai11rgbd}, containing 300 objects organized into 51 classes. 
In RGBD, we use 12 views for each 3D object as in~\cite{wei20viewgcn}. 
\end{itemize}

\begin{table}[t]\small
   \begin{center}
      \caption{Comparison of 3D shape recognition on ScanObjectNN.}
      \begin{tabular}{l l c c}
      \toprule
      \multirow{2}{*}{Method} & \multirow{2}{*}{Input} & Class Acc. & Inst. Acc.\\
      & & (\%) & (\%)\\
      \midrule
      PointNet++~\cite{han193D2SeqViews} & \multirow{4}{*}{Points} & 82.1 & 84.3 \\
      SpiderCNN~\cite{xu18spidercnn} & & 77.4 & 79.5\\
      PointCNN~\cite{li18pointcnn} & & -- & 85.5 \\  %
      DGCNN~\cite{wang19dgcnn} & & 84.0 & 86.2 \\\hdashline
      MVCNN-M & \multirow{7}{*}{Views} & 85.0 & 86.8 \\
      RotationNet~\cite{Kanezaki18rotationnet} &  & 84.9 & 86.9 \\
      CVR~\cite{wei21cvr} &  & 88.4 & 90.8 \\
      View-GCN~\cite{wei20viewgcn} &  & 88.3 & 90.4 \\
      View-GCN++~\cite{wei23viewgcn2} &  & {89.1} & 91.3 \\
      MVTN~\cite{ha19SeqViews2SeqLabels} &  & -- & 92.3 \\
      VointNet~\cite{hamdi23voint} &  & -- & {94.0} \\
      \midrule
      \textbf{VSFormer} & Views & \textbf{94.6} & \textbf{95.9}\\  %
      \bottomrule
      \end{tabular}
      \label{tab:recog_sonn}
   \end{center}
\end{table}

\begin{table}[t]\small
   \begin{center}
      \caption{Comparison of 3D shape recognition on RGBD.}
      \begin{tabular}{l c c}
      \toprule
      Method & \#Views & Inst. Acc. (\%)\\
      \midrule
      CFK~\cite{cheng15cfk} & $\geq$ 120 & 86.8\\
      MMDCNN~\cite{rahman17rgbdor} & $\geq$ 120 & 89.5\\
      MDSICNN~\cite{asif18amm} & $\geq$ 120 & 89.6\\
      MVCNN~\cite{su15mvcnn} & 12 & 86.1 \\
      RotationNet~\cite{Kanezaki18rotationnet} & 12 & 89.3 \\\hdashline
      View-GCN(ResNet18)~\cite{wei20viewgcn} & 12 & {94.3} \\
      View-GCN(ResNet50)~\cite{wei20viewgcn} & 12 & {93.9} \\
      \midrule
      \textbf{VSFormer}(ResNet18) & 12 & \textbf{98.4} \\
      \textbf{VSFormer}(ResNet50) & 12 & \textbf{95.6} \\
      \bottomrule
      \end{tabular}
      \label{tab:recog_rgbd}
   \end{center}
\end{table}

\noindent\textbf{Metrics.} 
Two evaluation metrics are computed for 3D shape recognition: mean class accuracy (Class Acc.) and instance accuracy (Inst. Acc.). 
We record the best results of these metrics during optimization. 

\noindent\textbf{Results.} 
Table~\ref{tab:recog_modelnet40} compares representative methods on ModelNet40 and these methods have different input formats: 
voxels, points and views. 
VSFormer achieves 98.9\% mean class accuracy and 98.8\% overall accuracy, surpassing the voxel- and point-based counterparts.  
Also, it sets new records in view-based methods. For example, compared to early works~\cite{su15mvcnn,su18mvcnn-new,yu18mhbn,feng18gvcnn,wang19rcpcnn}
that aggregate multi-view information independently by pooling or some variants,  
VSFormer exceeds their instance accuracies by 3.8\% at least. 
VSFormer also significantly improves the results of methods built on view sequence, such as RelationNet~\cite{yang19relationnet}, 
3D2SeqViews~\cite{han193D2SeqViews}, SeqViews2SeqLabels~\cite{ha19SeqViews2SeqLabels}, VERAM~\cite{chen19veram}. 
Methods defined on view graph and hyper-graph achieve decent performances~\cite{zhang19imhl,feng19hgnn,gao23hgnn+,wei20viewgcn,wei23viewgcn2} 
because of enhanced information flow among views. VSFormer still outreaches the strongest baseline of this category,  
increasing 2.4\% Class Acc. and 1.2\% Inst Acc. over View-GCN~\cite{wei20viewgcn}. 

Table~\ref{tab:recog_sonn} exhibits the evaluation results of various methods on the real-scan ScanObjectNN~\cite{uy19sonn} dataset. 
Apparently, VSFormer outreaches existing strong baselines in terms of class and instance recognition accuracies, 
leading View-GCN++~\cite{wei23viewgcn2} by 5.5\% class accuracy and VointNet~\cite{hamdi23voint} by 0.9\% instance accuracy. 
The results confirm the proposed method still works well when handling cluttered and occluded views. 

Table~\ref{tab:recog_rgbd} records the comparison with related work on the challenging RGBD~\cite{lai11rgbd} dataset. 
The dataset designs 10-fold cross-validation for multi-view 3D object recognition. 
We follow this setting and report the average instance accuracy of 10 folds. 
VSFormer shows consistent improvements over View-GCN under the same initializations. Especially, 
it gets 98.4\% accuracy, which is a 4.1\% absolute improvement over the runner-up, 
suggesting VSFormer can produce more expressive shape descriptors when dealing with challenging cases. 

\begin{table*}[t]\scriptsize
   \begin{center}
   \caption{Comparison of 3D shape retrieval on the normal version of ShapeNet Core55.}
   \begin{tabular}{l c c c c c c c c c c c}
      \toprule
      \multirow{2}{*}{Method} & \multicolumn{5}{c}{micro} & & \multicolumn{5}{c}{macro} \\
         \cline{2-6}                     \cline{8-12}
       & P@N & R@N & F1@N & mAP & NDCG & & P@N & R@N & F1@N & mAP & NDCG\\
      \midrule
      ZFDR & 53.5 & 25.6 & 28.2 & 19.9 & 33.0 & & 21.9 & 40.9 & 19.7 & 25.5 & 37.7 \\
      DeepVoxNet & 79.3 & 21.1 & 25.3 & 19.2 & 27.7 & & 59.8 & 28.3 & 25.8 & 23.2 & 33.7 \\
      DLAN & 81.8 & 68.9 & 71.2 & 66.3 & 76.2 & & 61.8 & 53.3 & 50.5 & 47.7 & 56.3 \\\hdashline
      GIFT~\cite{bai16gift} & 70.6 & 69.5 & 68.9 & 64.0 & 76.5 & & 44.4 & 53.1 & 45.4 & 44.7 & 54.8 \\
      Improved GIFT~\cite{bai17gift} & 78.6 & 77.3 & 76.7 & 72.2 & 82.7 & & 59.2 & 65.4 & 58.1 & 57.5 & 65.7 \\
      ReVGG & 76.5 & 80.3 & 77.2 & 74.9 & 82.8 & & 51.8 & 60.1 & 51.9 & 49.6 & 55.9 \\
      MVFusionNet & 74.3 & 67.7 & 69.2 & 62.2 & 73.2 & & 52.3 & 49.4 & 48.4 & 41.8 & 50.2 \\
      CM-VGG5-6DB & 41.8 & 71.7 & 47.9 & 54.0 & 65.4 & & 12.2 & 66.7 & 16.6 & 33.9 & 40.4 \\
      MVCNN~\cite{su15mvcnn} & 77.0 & 77.0 & 76.4 & 73.5 & 81.5 & & 57.1 & 62.5 & 57.5 & 56.6 & 64.0 \\
      RotationNet~\cite{Kanezaki18rotationnet} & 81.0 & 80.1 & 79.8 & 77.2 & 86.5 & & 60.2 & 63.9 & 59.0 & 58.3 & 65.6 \\
      View-GCN~\cite{wei20viewgcn} & 81.8 & {80.9} & {80.6} & 78.4 & {85.2} & & {62.9} & 65.2 & {61.1} & {60.2} & {66.5} \\
      View-GCN++~\cite{wei23viewgcn2} & 81.2 & 79.9 & 80.0 & {77.5} & 83.9 & & 61.2 & {65.8} & {61.1} & 59.0 & 63.8 \\
      MVPNet~\cite{xu24mvpnet} & 81.2 & 81.1 & 80.8 & 78.5 & 86.3 & & 61.5 & 66.9 & 62.5 & 61.2 & 66.5 \\
      MRVA-Net~\cite{lin23mrvanet} & 80.0 & \textbf{83.0} & 80.8 & 79.2 & \textbf{87.3} & & 61.2 & \textbf{69.8} & 62.3 & 63.1 & 69.6 \\
      \midrule
      $\textbf{VSFormer}$ & \textbf{82.3} & 82.6 & \textbf{82.0} & \textbf{79.3} & 82.4 & & \textbf{67.0} & 69.7 & \textbf{66.7} & \textbf{64.2} & \textbf{70.4} \\ 
      \bottomrule
   \end{tabular}
   \label{tab:ret_shrec17}
   \end{center}
\end{table*}
\begin{table*}[t]\scriptsize
   \centering
   \caption{Ablation Study: the architecture of Encoder.}
   \begin{tabular}{l r r r r r r r r r r r r r }
      \toprule
      \#Blocks & 2 & 2 & 2 & 2 & 4 & 4 & 4 & 4 & 6 & 6 & 6 & 6 \\ 
      \#Heads & 6 & 8 & 6 & 8 & 6 & 8 & 6 & 8 & 6 & 8 & 6 & 8 \\
      Ratio$_{mlp}$ & 2 & 2 & 4 & 4 & 2 & 2 & 4 & 4 & 2 & 2 & 4 & 4 \\
      Dim$_{view}$ & 384 & 512 & 384 & 512 & 384 & 512 & 384 & 512 & 384 & 512 & 384 & 512 \\
      \midrule
      \#Params (M) & 2.7 & 4.8 & 3.9 & 6.9 & 5.0 & 9.0 & 7.4 & 13.2 & 7.4 & 13.2 & 11.0 & 19.5\\
      \midrule
      ModelNet40 \\
      \quad Class Acc. (\%) & 98.8 & 98.7 & 98.4 & 97.2 & 97.4 & {98.9} & \textbf{99.1} & 98.2 & 98.7 & 98.2 & 98.4 & 98.1 \\
      \quad Inst. Acc. (\%) & \textbf{99.0} & {98.8} & 98.5 & 98.1 & 97.6 & {98.8} & 98.5 & 98.5 & 98.3 & 98.3 & 98.1 & 98.3 \\
      \bottomrule
   \end{tabular}
   \label{tab:encoder_arch}
\end{table*}
\subsection{3D Shape Retrieval}
\label{subsec:exp_retrieval}

\noindent\textbf{Datasets.}
3D shape retrieval aims to find a rank list of shapes most relevant to the query in a given dataset. 
We conduct this task on ShapeNet Core55~\cite{shapenet2015,savva17shrec17}. The dataset is split into train/val/test 
set and there are 35764, 5133 and 10265 meshes, respectively. 
20 views are rendered for each mesh as in~\cite{wei20viewgcn} and we explain the procedure
in this blog\footnote{\url{https://auniquesun.github.io/2023-01-15-shapenetcore55-rendering/}}. 
ShapeNet Core55 has two rendered versions (normal and perturbed) 
and we report results on the normal one as in previous work~\cite{Kanezaki18rotationnet,ma19learning,wei20viewgcn}. 

\noindent\textbf{Metrics.}
According to the SHREC'17 benchmark~\cite{savva17shrec17}, the rank list is evaluated based on the ground truth 
category and subcategory. If a retrieved shape in a rank list has the same category as the query, it is positive. Otherwise, it is negative. 
The evaluation metrics include micro and macro P@N, R@N, F1@N, mAP and NDCG. 
P@N and R@N mean the precision and recall when the length of the returned rank list is N (1,000 by default).
NDCG that represents normalized discounted cumulative gain is a measure of ranking quality.
Please refer to \cite{savva17shrec17} for more details about the metrics. 

\noindent\textbf{Retrieval.} 
We generate the rank list for each query shape in two steps. 
First, VSFormer is trained to recognize the shape categories in ShapeNet Core55~\cite{shapenet2015}. 
We retrieve shapes that have the same predicted class as the query $\mathcal{Q}$ and rank the retrieved shapes according to 
class probabilities in descending order, resulting in L$_1$.
Second, we train another VSFormer to recognize the shape subcategories of ShapeNet Core55~\cite{shapenet2015}, 
then re-rank L$_1$ to ensure shapes that 
have same predicted subcategory as the query $\mathcal{Q}$ rank before shapes that are not in same subcategory with $\mathcal{Q}$
and keep the remaining unchanged, 
resulting in L$_2$, which is regarded as the final rank list for the query $\mathcal{Q}$. 

\noindent\textbf{Results.} 
VSFormer is compared with the methods that report results on SHREC'17 benchmark~\cite{savva17shrec17}, 
shown in Table~\ref{tab:ret_shrec17}. 
The methods in the first three rows use voxels as inputs, while the remaining ones exploit views. 
The overall performances of view-based methods are better than voxel-based ones.  
Previously, MRVA-Net achieved state-of-the-art results by extracting multi-range view features and fusing the features with ViT. 
But experiments show VSFormer goes beyond MRVA-Net in 7 out 10 metrics, including 
micro P@N, F1@N and mAP as well as macro P@N, F1@N, mAP and NDCG. 
In particular, we achieve 2.3\% and 5.8\% absolute improvements for micro and macro P@N over MRVA-Net. 
On the other hand, the Transition module implemented as a concatenation of max and mean pooling 
will inevitably lose a bunch of grasped correlations. The model can be confused by the highly similar 3D shapes 
in appearance and suffers from poor rankings for the shapes, which results in lower micro NDCG.

\subsection{Ablation Studies}
\label{subsec:ablation_studies}
We conducted controlled experiments to verify the choices of different modules in VSFormer design, 
analyze the impact of patch-level correlations and the number of views. 
The used dataset is ModelNet40. 

\subsubsection{Encoder}
\noindent\textbf{The Architecture of Encoder.} 
We provide ablations to justify the design choices of Encoder.
The controlled variables of Encoder are the number of attention blocks (\#Blocks), the number of attention heads in MSA (\#Heads), 
the widening ratio of MLP hidden layer (Ratio$_{mlp}$) and the dimension of the view representations (Dim$_{view}$). 
The mean class acc. and instance acc. of VSFormer with different encoder structures are compared in Table~\ref{tab:encoder_arch}. 
All design variants show high-level performances and surpass the existing state of the art. 
Surprisingly, the encoder consisting of only 2 attention blocks can facilitate VSFormer to achieve 99.0\% overall accuracy. 
The results are in line with expectations as the size of a view set is relatively small thus, it is unnecessary to 
design a heavy encoder. At the same time, it is inspiring that a shallow encoder can enrich and grasp the pairwise and 
higher-order correlations of elements in the view set well. Finally, we select the design that takes \emph{the second place} in both mean class and 
instance accuracy, namely \#Blocks = 4, \#Heads = 8, Ratio$_{mlp}$ = 2 and Dim$_{view}$ = 512. 

\noindent\textbf{Performance Gains Delivered by Our Encoder.} 
We investigate the performance gains delivered by the devised view set encoder. 
There are two settings. 1) The initializer is individually trained to recognize 3D shapes. 
2) The devised encoder is appended upon the pre-trained initializer to further capture the feature interactions among views. 
Table~\ref{tab:performance_gains_by_encoder} compares 
different configurations described above.
Notable performance gains are obtained over different initializers. 
For example, by appending only 2 attention blocks (4.8M \#Params) on the AlexNet initializer, 
our model achieves 18.3\% and 13.7\% absolute improvements for mean class accuracy and instance accuracy.
\begin{table}[ht]
   \centering
   \caption{Ablation study: the performance gains brought by the devised encoder.}  %
   \begin{tabular}{l l l l}
      \toprule
      Module & \#Params(M) & Class Acc.(\%) & Inst. Acc.(\%) \\
      \midrule
      AlexNet \\
      $\quad$w/o encoder & {42.3} & {80.6} & {85.1} \\  %
      $\quad$w/ encoder & 47.1$\small{\uparrow{4.8}}$ & \textbf{98.9}$\small{\uparrow{18.3}}$ & \textbf{98.8}$\small{\uparrow{13.7}}$ \\
      \midrule
      ResNet18 \\
      $\quad$w/o encoder & {11.2} & {88.7} & {91.8} \\ %
      $\quad$w/ encoder & 16.0$\small{\uparrow{4.8}}$ & \textbf{96.7}$\small{\uparrow{8.0}}$ & \textbf{97.6}$\small{\uparrow{5.8}}$ \\  %
      \bottomrule
   \end{tabular}
   \label{tab:performance_gains_by_encoder}
\end{table}

\noindent\textbf{Position Encoding.}
According to the view set perspective, 
VSFormer should be unaware of the order of elements in the view set, so we remove the position 
encoding from the devised encoder. We examine this design in Table~\ref{tab:ablate_pos_enc}.
The results show 
if learnable position embeddings are forcibly injected into the initialized view features to make 
the model position-aware, the performance will be hindered, dropping by 0.5\% for class accuracy
and 0.3\% for overall accuracy. 

\noindent\textbf{Class Token.} Unlike standard Transformer~\cite{vaswani17transformer}, 
the proposed method does not insert the class token into the inputs since 
it is irrelevant to the target of capturing the correlations among views in the set. 
This claim is supported by the results in Table~\ref{tab:ablate_pos_enc}, which shows that 
inserting the class token decreases recognition accuracies. 
\begin{table}[ht]
	\begin{center}
      \caption{Ablation study: position encoding and class token.}
		\begin{tabular}{l l l}
			\toprule
			Variants & Class Acc. (\%) & Inst. Acc. (\%) \\
			\midrule
			w/ pos. enc. & {98.4} & {98.5}\\
			w/o pos. enc. & \textbf{98.9}$\small{\uparrow{0.5}}$ & \textbf{98.8}$\small{\uparrow{0.3}}$\\
			\midrule
			w/ cls. token & {98.8} & {98.5} \\
			w/o cls. token & \textbf{98.9}$\small{\uparrow{0.1}}$ & \textbf{98.8}$\small{\uparrow{0.3}}$\\
			\bottomrule
		\end{tabular}
		\label{tab:ablate_pos_enc}
	\end{center}
\end{table}

\noindent\textbf{Number of Attention Blocks.} 
In VSFormer, the number of attention blocks in Encoder is considerably compressed because 
the size of a view set is relatively small and it is unnecessary to deploy a deeper encoder to model 
the interactions between the views in the set. The results in Table~\ref{tab:ablate_num_attn_blocks}
demonstrate the encoder can be highly lightweight, as light as 2 attention blocks, but with 98.8\% overall accuracy 
that exceeds all existing methods. The results also indicate 
increasing the attention blocks does not receive gains but additional parameters and overheads. 
\begin{table}[ht]
	\begin{center}
      \caption{Ablation study: number of attention blocks.}  %
		\begin{tabular}{l r l}
			\toprule
			Module & \#Params (M) & Inst. Acc. (\%) \\
			\midrule
			AlexNet & 42.3 & 85.1 \\
			\quad + 2 Attn. Blocks & \textbf{4.8} & \textbf{98.8}$\small{\uparrow{13.7}}$ \\
			\quad + 4 Attn. Blocks & {9.0} & \textbf{98.8}$\small{\uparrow{13.7}}$ \\
			\quad + 6 Attn. Blocks & 13.2 & {98.3}$\small{\uparrow{13.2}}$ \\
			\bottomrule
		\end{tabular}
		\label{tab:ablate_num_attn_blocks}
	\end{center}
\end{table}

\subsubsection{Initializer}
We explore different means to initialize view representations, 
including shallow convolution operations and lightweight CNNs. 
The idea of shallow convolution operation is inspired by the image patch projection (1x1 Conv) in ViT~\cite{dosovitskiy21vit} and 
the specific configurations are explained in Table~\ref{tab:init_shallow_convs}. 
Table~\ref{tab:ablate_view_init} compares their recognition accuracies. 
We observe that initializations by 1- and 2-layer convolution operations do not yield satisfactory
results. Instead, 
lightweight CNNs work well, especially when receiving the initialized features from AlexNet and jointly optimizing with other modules, 
VSFormer reaches 98.9\% class accuracy and 98.8\% overall accuracy, both are new records on ModelNet40. 
By default, AlexNet serves as the Initializer module.
\begin{table}[ht]
	\begin{center}
      \caption{Ablation study: choices for Initializer.}
		\begin{tabular}{l r c c}
			\toprule
			\multirow{2}{*}{Initializer} & \#Params &Class Acc.& Inst. Acc. \\
			& (M) & (\%) & (\%) \\
			\midrule
			1-layer Conv & 102.8 & 90.1 & 92.5 \\  %
			2-layer Conv & 12.9 & 88.9 & 93.7 \\\hdashline  %
			alexnet & 42.3 & \textbf{98.9} & \textbf{98.8} \\ %
			resnet18 & 11.2 & 96.7 & {97.6}\\
			resnet34 & 21.3 & {96.9} & 97.1 \\
			\bottomrule
		\end{tabular}
		\label{tab:ablate_view_init}
	\end{center}
\end{table}

\subsubsection{Transition}
We investigate three kinds of operations for the Transition module. The results are reported 
in Table~\ref{tab:ablate_transition_module}. We find the simple pooling operations (Max and Mean) can work well (98.0+\% Acc.) and
both outreach the performances of the previous state of the art. By concatenating the outputs of max and mean pooling, 
the optimization is more stable and the overall accuracy is lifted to 98.8\%. 
It is worth noting that the same pooling operations are adopted by MVCNN~\cite{su15mvcnn} and its variants~\cite{su18mvcnn-new,feng18gvcnn,yu18mhbn,wang19rcpcnn,yu21mhbn++}, 
but their accuracies are up to 95.0\%, implying that the view set descriptors learned by our encoder are more informative. 
\begin{table}[ht]
	\begin{center}
      \caption{Ablation study: choices for Transition.}
		\begin{tabular}{l c c}
			\toprule
			Transition & Class Acc. (\%) & Inst. Acc. (\%) \\
			\midrule
			Max pooling & \textbf{99.1} & {98.5} \\
			Mean pooling & 98.5 & {98.5} \\
			Concat(Max\&Mean) & {98.9} & \textbf{98.8} \\
			\bottomrule
		\end{tabular}
		\label{tab:ablate_transition_module}
	\end{center}
\end{table}

\subsubsection{Decoder}
The decoder projects the view set descriptor to a shape category distribution. 
The choices for the decoder are compared in Table~\ref{tab:ablate_decoder}. 
VSFormer with a decoder of a single Linear can recognize 3D shapes at 98.1\% instance accuracy, 
which outperforms all existing methods and again, reflects the summarized view set descriptor is highly discriminative. 
The advantage is enlarged when the decoder is deepened to a 2-layer MLP. However, further tests show 
it is unnecessary to exploit deeper transformations. %
\begin{table}[ht]
	\begin{center}
      \caption{Ablation study: choices for Decoder.}
		\begin{tabular}{l c c}
			\toprule
			Decoder & Class Acc. (\%) & Inst. Acc. (\%) \\
			\midrule
			1-layer & 97.9 & 98.1 \\   %
			2-layer & \textbf{98.9} & \textbf{98.8} \\   %
			3-layer & {98.5} & {98.5} \\   %
			\bottomrule
		\end{tabular}
		\label{tab:ablate_decoder}
	\end{center}
\end{table}

\subsubsection{Effect of the Patch-level Feature Correlations}
Some other methods, such as MHBN~\cite{yu18mhbn,yu21mhbn++}, MVT~\cite{chen21mvt}, CarNet~\cite{xu21carnet}, 
also consider patch-level interactions. They want 
to enhance multi-view information flow by integrating patch-level features. In this work, we examine the effect of 
patch-level feature correlations by injecting them into each attention block of the encoder. The results in 
Table~\ref{tab:patch_level_correlations} show injecting patch-level features is redundant and unnecessary. 
A major reason is the view-level correlations are already well understood by our model for 3d shape analysis. 
Many fine-grained patches are similar in different views and lack the sense of overall shape, 
some of them are even blank backgrounds, thus contribute little to the target task. 
However, no matter with or without the patch-level correlations, 
VSFormer maintains high-level performances (98.1\% class and inst. accuracies) and surpasses all existing models. 
\begin{table}[ht]
   \begin{center}
   \caption{Ablation study: effect of the patch-level correlations.}
      \begin{tabular}{l l l}
      \toprule
      Variants & Class Acc. (\%) & Inst. Acc. (\%)\\
      \midrule
      w/ patch & {98.1} & {98.1}\\    %
      w/o patch & \textbf{98.9}$\small{\uparrow{0.8}}$ & \textbf{98.8}$\small{\uparrow{0.7}}$\\ %
      \bottomrule
      \end{tabular}
   \label{tab:patch_level_correlations}
   \end{center}
\end{table}

\subsubsection{Effect of the Number of Views}
We investigate the effect of the number of views on the recognition performance, shown in Table~\ref{tab:ablate_num_views}. 
There are up to 20 views for each 3D shape and $M$ views are randomly selected for each shape during training and evaluation, 
where $M \in$ \{1, 4, 8, 12, 16, 20\}. 
When $M=1$, the problem is equivalent to single-view object recognition, so there is no interaction among views. In this case,
a lightweight ResNet18~\cite{he16resnet} is trained for recognition and it achieves 89.0\% mean class accuracy and 91.8\% instance accuracy. 
When the number of views increases, the performances are quickly improved. For instance, after aggregating the correlations from 4 views, 
VSFormer lifts 8.4\% and 5.3\% absolute points in class and instance accuracy, respectively. 
But exploiting more views does not necessarily lead to better accuracy. 
The 8-view VSFormer reaches 98.0\% class accuracy and 98.8\% overall accuracy, outperforming 12- and 16-view versions. 
The performance is optimal when exploiting all 20 views and we choose this version to compare with other view-based methods. 
\begin{table}[ht]
   \begin{center}
   \caption{Ablation study: effect of the number of views.}  %
   \begin{tabular}{l r r r r r r}
      \toprule
      \#Views & 1 & 4 & 8 & 12 & 16 & 20 \\
      \midrule
      Class Acc. (\%) & 89.0 & 97.4 & {98.0} & 97.5 & 97.7 & \textbf{98.9} \\
      \midrule
      Inst. Acc. (\%) & 91.8 & 97.1 & \textbf{98.8} & 97.6 & {98.3} & \textbf{98.8} \\
      \bottomrule
   \end{tabular}
   \label{tab:ablate_num_views}
   \end{center}
\end{table}

\section{Visualization}
\label{sec:vis}

This section exhibits various visualizations of the predictions and intermediate results 
given by VSFormer, which are helpful for having a better understanding of our method.

\begin{figure}[ht]
   \centering
   \includegraphics[width=\linewidth]{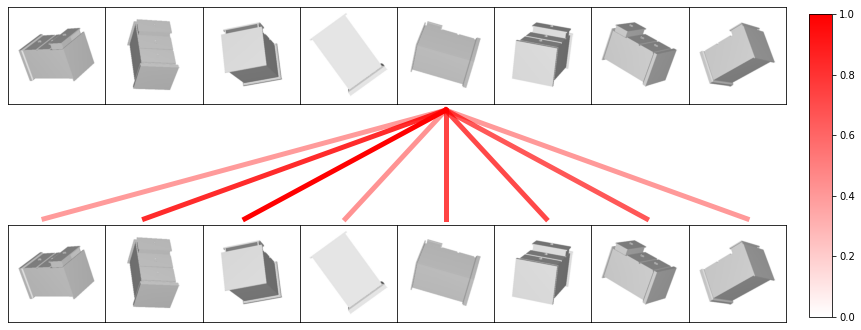}
   \caption{Visualization of multi-view attention of 8 views of a nightstand in colored lines.}
   \label{fig:8nightstands_attn_lines}
\end{figure}

\noindent\textbf{Multi-view Attention in Colored Lines.}
We randomly select a 3D shape that is a nightstand, then visualize the multi-view correlations of eight views of this 
shape, referring to Figure~\ref{fig:8nightstands_attn_lines}. The correlations are represented by the attention scores emitted 
by the last attention block of VSFormer. The scores are normalized to map to the color bar on the far right of the figure. 
Our model distributes more weights to the 2nd, 3rd and 6th views from the 5th. The results seem reasonable since these views 
are more discriminative according to visual appearances.

\begin{figure}[ht]
   \begin{center}
      \includegraphics[width=\linewidth]{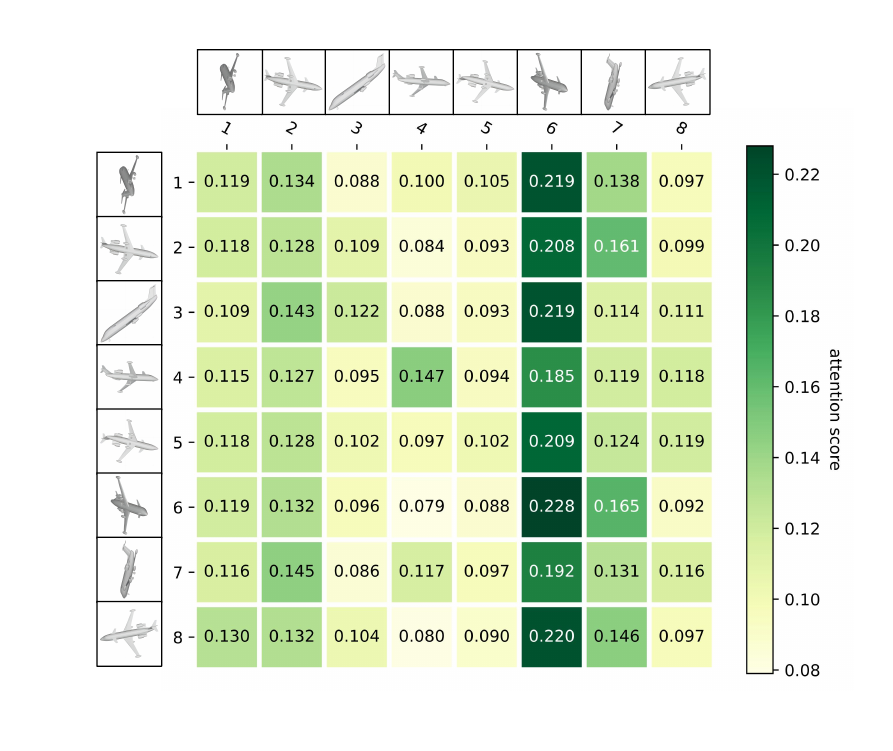}
   \end{center}
   \caption{Visualization of the attention scores for 8 views of a 3D airplane.}
   \label{fig:8airplanes_attn_map}
\end{figure}

\noindent\textbf{Multi-view Attention Map.} 
For better understanding, we visualize the attention map of eight views of a 3D airplane in Figure~\ref{fig:8airplanes_attn_map}. 
The attention scores are taken from the outputs of 
the last attention block of our model. 
We normalize the attention scores so that each score is rounded to three decimal digits and the sum of the scores for each row is equivalent to 1. 
Based on the 3rd or 7th view, one may not be confident that the shape is a airplane, and our model assigns relatively small weights to them. 
Instead, the map indicates the 6th view is representative since it receives more attention from other views. 
On the other hand, we can manually infer the 6th view is representative based on the visual appearances of these views. 
The results reflect that the proposed model can adaptively capture the multi-view correlations and allocate different views with reasonable weights 
for recognition.

\begin{figure}[ht]
	\centering
   \begin{subfigure}{0.33\linewidth}
      \includegraphics[width=\linewidth]{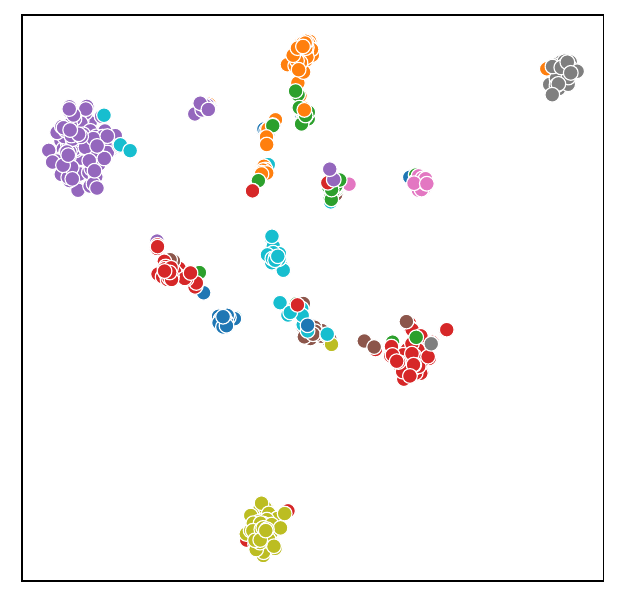}
      \caption{SONN}
      \label{fig:sonn_feats}
   \end{subfigure}%
   \begin{subfigure}{0.33\linewidth}
      \includegraphics[width=\linewidth]{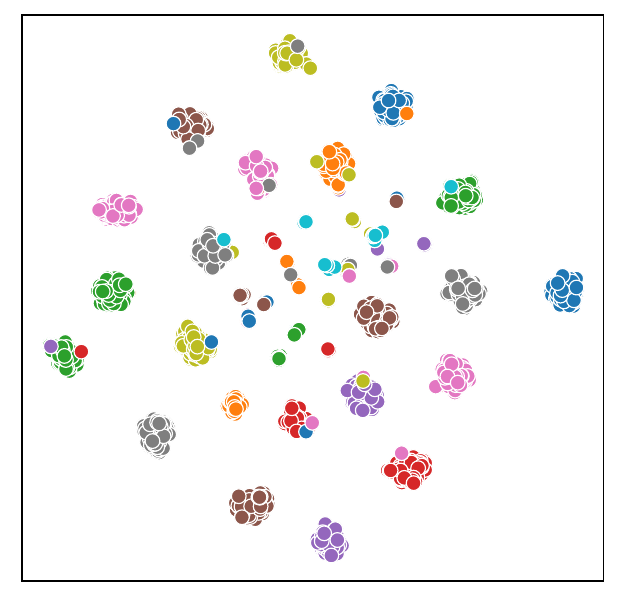}
      \caption{MN40}
      \label{fig:mn40_feats}
   \end{subfigure}%
   \begin{subfigure}{0.33\linewidth}
      \includegraphics[width=\linewidth]{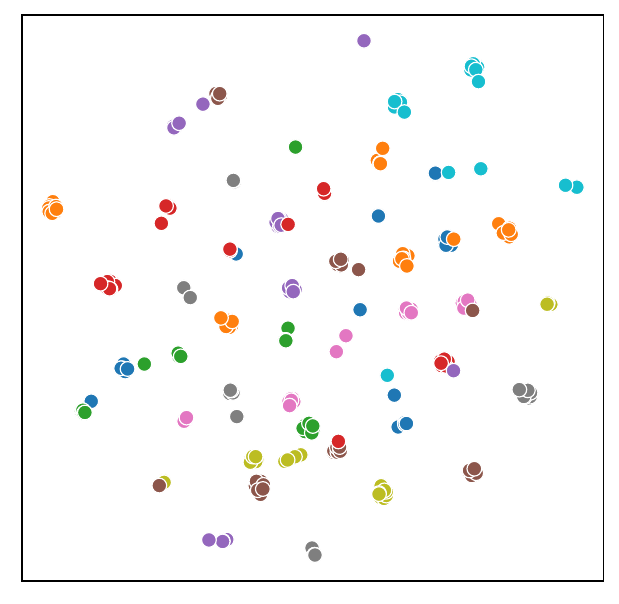}
      \caption{RGBD}
      \label{fig:rgbd_feats}
   \end{subfigure}
	\caption{Visualization of 3D shape \textbf{feature distribution} on (a) ScanObjectNN (SONN) of 15 classes 
   (b) ModelNet40 (MN40) of 40 classes (c) RGBD of 51 classes.}
	\label{fig:recog_feats_visual}
\end{figure}

\noindent\textbf{3D Shape Recognition.} 
We visualize the feature distribution for different shape categories on ScanObjectNN, ModelNet40 and RGBD using t-SNE~\cite{tsne}, 
shown in Figure~\ref{fig:recog_feats_visual}. 
It shows different shape categories in different datasets are successfully distinguished by the proposed method, 
demonstrating that VSFormer understands multi-view information well by 
explicitly modeling the correlations for all view pairs in the view set. 

\begin{figure}[ht]
	\begin{center}
		\includegraphics[width=\linewidth]{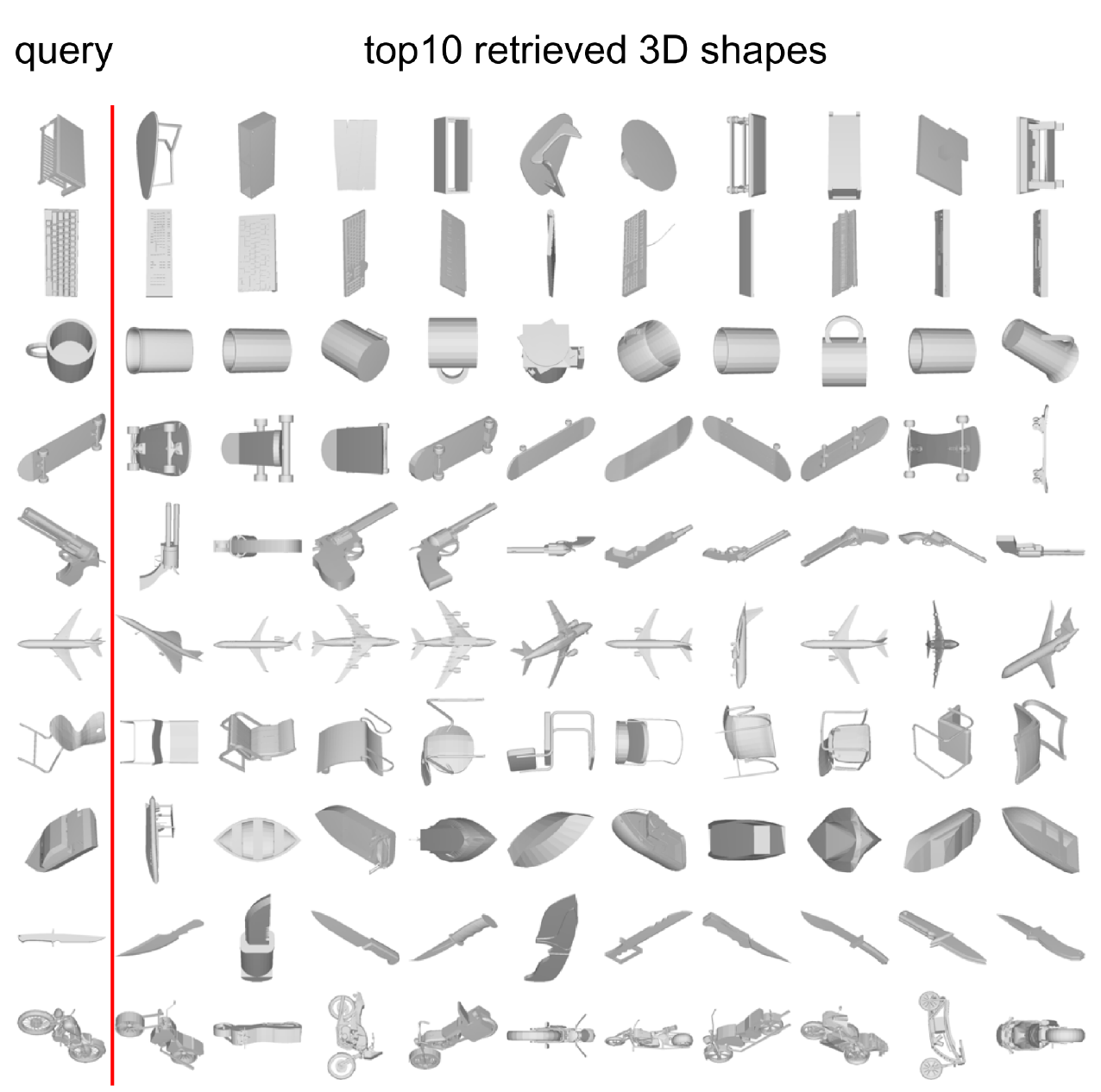}
	\end{center}
	\caption{Visualization of the top 10 retrieved results for each query shape.}
	\label{fig:shape_retrieval_example}
\end{figure}

\noindent\textbf{3D Shape Retrieval.} 
We visualize the top 10 retrieved shapes for 10 typical queries in Figure~\ref{fig:shape_retrieval_example}. 
The retrieval happens in the ShapeNet Core55 validation set. Each retrieved shape is represented by its random view. 
We observe the top 10 results are 100\% relevant to the query, which means they belong to the same category. 
The 5th retrieved shape in the 3rd row may be confusing. But when observing other views of this shape in ShapeNet,
it is also a cup. 

\section{Additional Analysis}
\label{sec:additional_analysis}
We provide additional analysis of the proposed approach, including network training, inference speed, 
and auxiliary observations in the initializer module. 

\subsection{Network Training}
The following settings will verify the adopted 2-stage training strategy and excellent learning efficiency
of VSFormer. 

\begin{figure*}[t]
   \centering
   \begin{subfigure}{0.32\linewidth}
      \includegraphics[width=\linewidth]{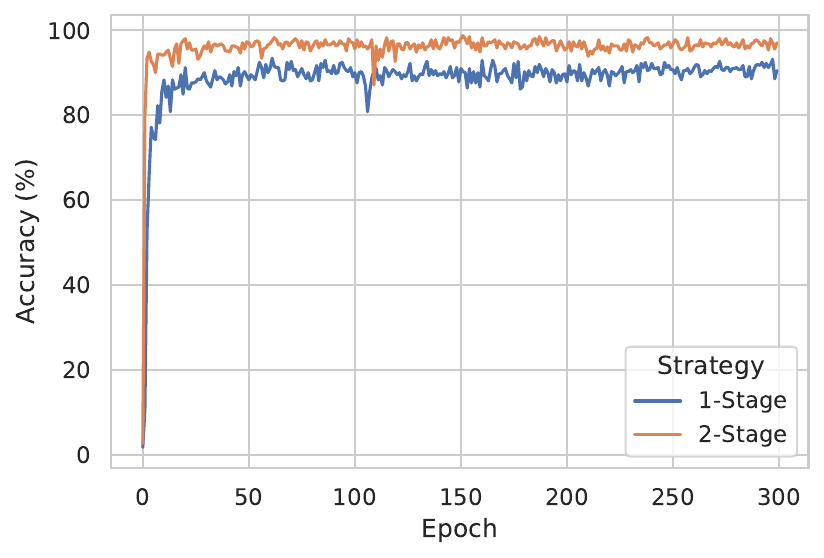}
      \caption{Training strategy.}
      \label{fig:VSFormer_training_strategy}
   \end{subfigure}%
   \begin{subfigure}{0.33\linewidth}
      \includegraphics[width=\linewidth]{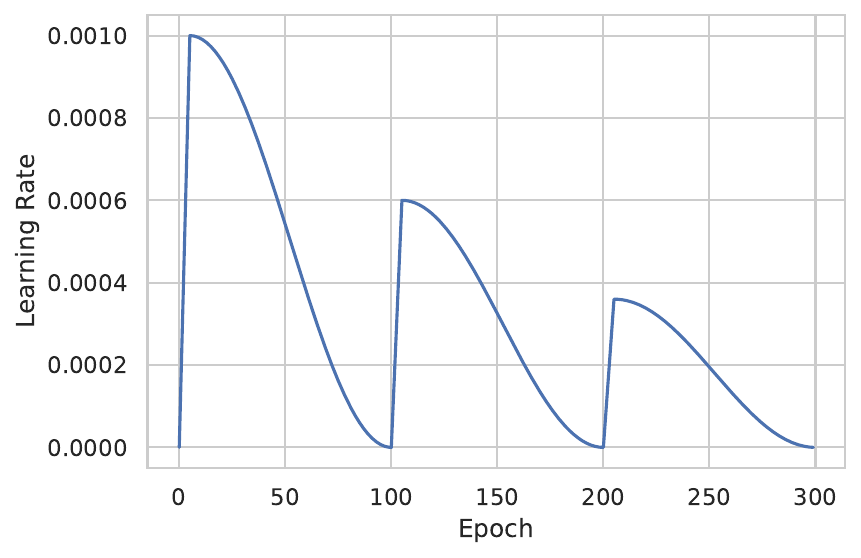}
      \caption{Learning rate curve.}
      \label{fig:VSFormer_lr}
   \end{subfigure}%
   \begin{subfigure}{0.32\linewidth}
      \includegraphics[width=\linewidth]{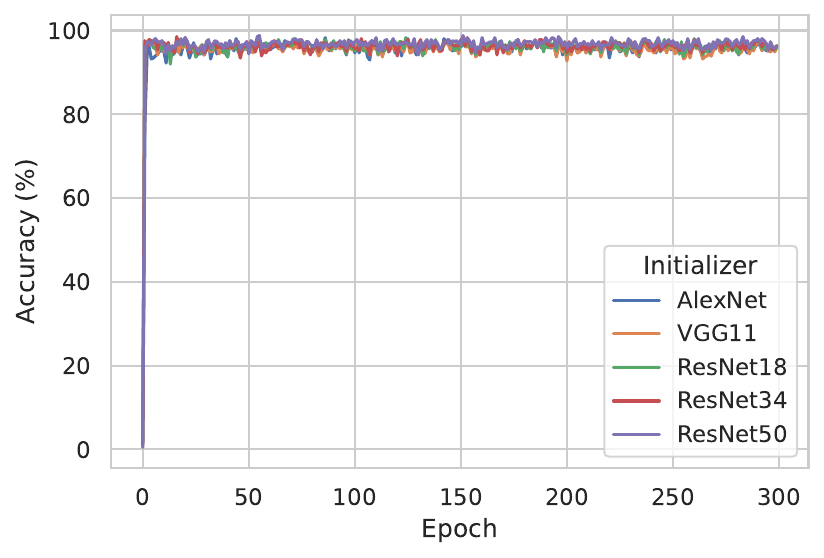}
      \caption{Learning efficiency.}
      \label{fig:learning_efficiency}
   \end{subfigure}
   \caption{(a) Comparison of instance accuracy using 1-stage and 2-stage optimization on ModelNet40.
   (b) The learning rate curve of AdamW for VSFormer.
   (c) Learning efficiency of the VSFormer variants using different initializers.}
\end{figure*}

\noindent\textbf{Optimization Strategy.}
We compare the effectiveness of 1-stage and 2-stage optimization on ModelNet40. For 2-stage optimization, Initializer is trained 
on the dataset individually, then the pre-trained weights of Initializer are loaded into VSFormer to be jointly optimized with
other modules. The 1-stage optimization means VSFormer learns in an end-to-end way and all parameters are randomly initialized. 
Figure~\ref{fig:VSFormer_training_strategy} shows the recognition accuracy achieved by 2-stage optimization is significantly 
better than that of 1-stage training. 
The results demonstrate that VSFormer receives gains from the well-initialized view representations 
provided by the first stage. 

\noindent\textbf{Learning Efficiency.}
We explore the learning efficiency of VSFormer by freezing the weights of the pre-trained Initializer. Figure~\ref{fig:learning_efficiency} 
displays the recognition accuracy curves of the VSFormer variants with different initializers on ModelNet40 during training. 
Regardless of Initializer used, all variants' performances soared after a short training and approached the highest. 
For instance, VSFormer with ResNet34 Initializer reaches 97.6\% instance accuracy after \emph{only 2 epochs}, while View-GCN achieves the
same performance with 7.5x longer optimization. The results reflect the proposed method has higher learning efficiency than the previous state of the art. 

\subsection{Inference Speed}
Table~\ref{tab:methods_params_and_inference_speed} compares the number of parameters, 
inference speed and recognition accuracy of different models. 
The experiments take place on ModelNet40 using a 2080Ti GPU. 
The inference speed counts the number of 3D objects each model processes per second (obj/s). 
Only the forward pass that outputs the prediction is considered, 
not including data preparation and loss computation. 
Note that each object corresponds to 20 views. 
OA is short for overall accuracy and mAcc refers to mean class accuracy. 

We select MVCNN and View-GCN for comparison. 
The choice considers the availability and usability of the official code. 
The former is a pioneer work in the field and the latter was the previous SOTA method. 
MVCNN has the least \#Params and runs much faster than View-GCN and VSFormer. 
View-GCN has the most \#Params and significantly improves recognition 
accuracy over MVCNN.
But when combining these metrics as a whole, VSFormer demonstrates consistent advantages under different initializers, 
achieving efficient inference speed and best recognition results. 
The proposed method is 3.8-6.4x faster than View-GCN when employing different initializers. 
Note that a large portion of parameters in VSFormer are attributed to the initializer, and the view set encoder is highly lightweight. 
For example, our encoder contains only 9.0M parameters (51.3M-42.3M), compared to 42.3M in AlexNet. 
Readers may notice that VSFormer with AlexNet (51.3M \#Params) is faster than that with ResNet18 (20.1M \#Params). 
It is due to AlexNet having fewer multiply-add operations than ResNet18, $e.g.$, 0.72 vs. 1.82 in GFLOPS. 

\begin{table}[ht]\scriptsize
   \begin{center}
      \caption{Comparison of the number of parameters, inference speed and accuracy of different methods.}
      \begin{tabular}{l c r r c c}
      \toprule
      \multirow{2}{*}{Method} & \multirow{2}{*}{Initializer} & \#Params & Speed & OA & mAcc\\
       & & (M) & (obj/s) & (\%) & (\%)\\
      \midrule
      MVCNN~\cite{su15mvcnn} & \multirow{3}{*}{ResNet18} & \textbf{11.2} & \textbf{1621.3} & 91.7 & 87.3 \\ 
      View-GCN~\cite{wei20viewgcn} &  & 33.9 & 131.8 & 96.1 & 96.1 \\ 
      \textbf{VSFormer} & & 20.1 & 979.4 & \textbf{97.1} & \textbf{96.4} \\\hdashline %
      MVCNN~\cite{su15mvcnn} & \multirow{3}{*}{AlexNet} & \textbf{42.3} & \textbf{2710.0} & 90.0 & 86.4 \\
      View-GCN~\cite{wei20viewgcn} &  & 84.6 & 325.7 & 91.5 & 88.9 \\ 
      \textbf{VSFormer} & & 51.3 & 1562.5 & \textbf{97.8} & \textbf{97.5} \\ %
      \bottomrule
      \end{tabular}
      \label{tab:methods_params_and_inference_speed}
   \end{center}
\end{table}

\subsection{Bad Case Analysis}
\begin{figure}
    \centering
    \subfloat[Incorrect predictions for multi-view shape recognition on ModelNet40]{
        \includegraphics[width=\linewidth]{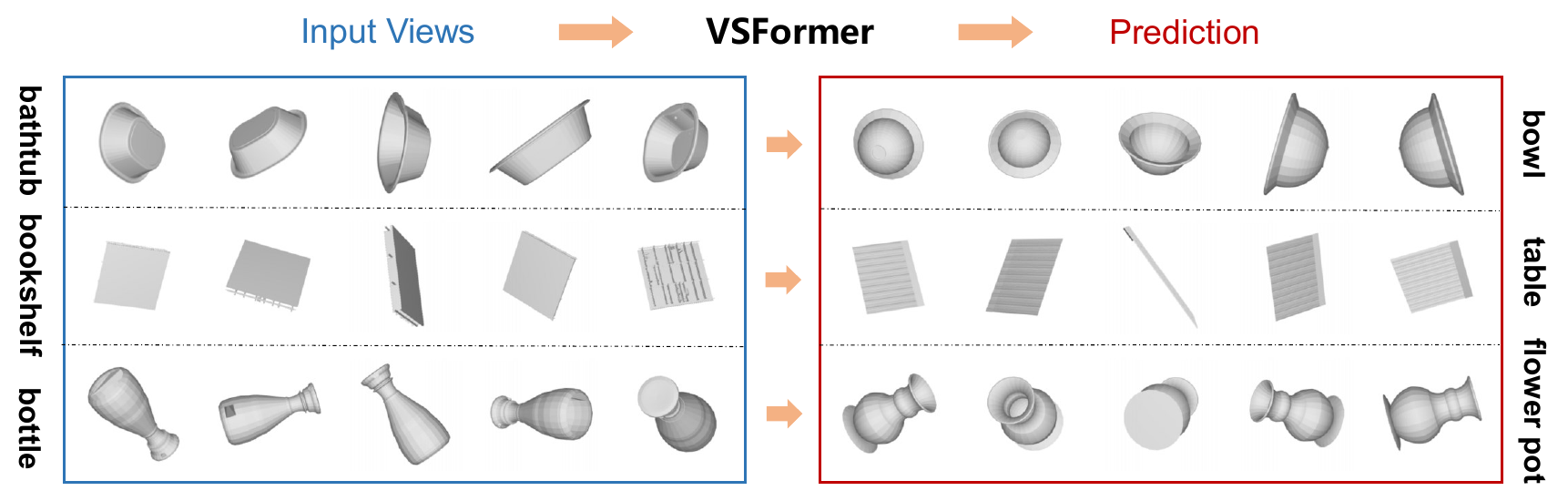}
        \label{fig:mn40_recognition_bad_cases}
    }\\
    \subfloat[Incorrect predictions for multi-view shape retrieval on SHREC'17]{
        \includegraphics[width=\linewidth]{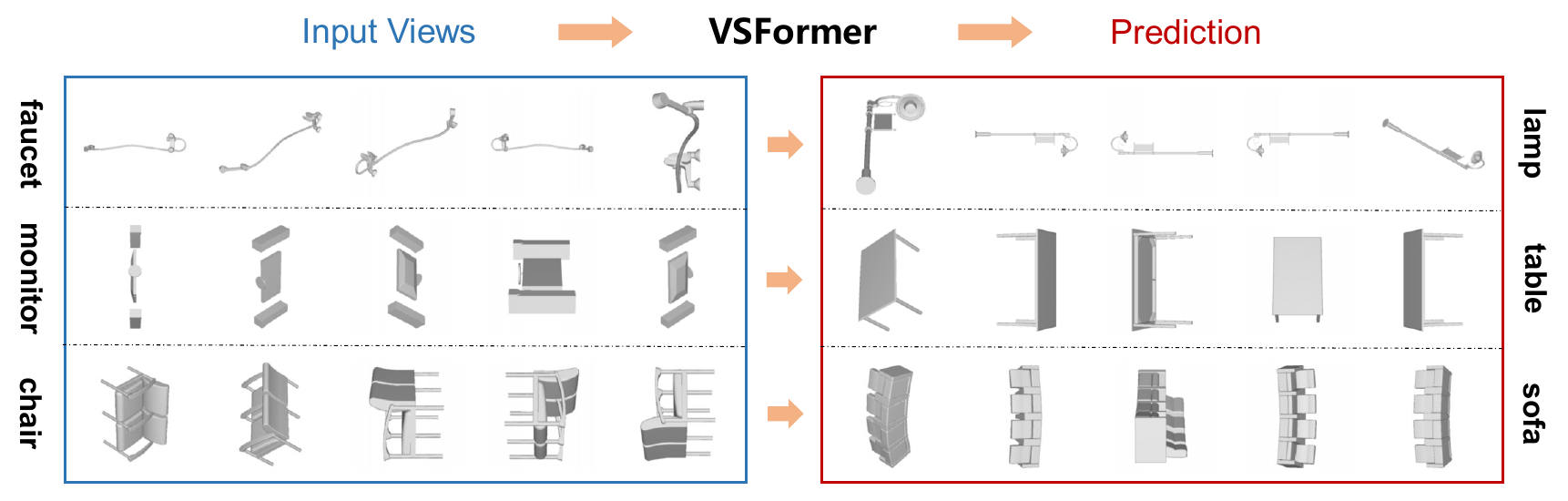}
        \label{fig:shrec17_retrieval_bad_cases}
    }
    \caption{Bad case analysis of the proposed model for the multi-view recognition and retrieval tasks.}
    \label{fig:recognition_retrieval_bad_cases}
\end{figure}

Here we carry out a bad case analysis of the proposed model. The study covers recognition and retrieval tasks on different datasets, including ModelNet40 and SHREC'17. The input views come from the corresponding test set and the incorrect predictions of our model are visualized in Figure~\ref{fig:recognition_retrieval_bad_cases}. 
For the recognition task, the model is confused by shapes with highly similar appearances, resulting in incorrect outputs on occasion. 
As the subfigure~\ref{fig:mn40_recognition_bad_cases} displays, our model predicts the \emph{bathtub} views as the \emph{bowl} category, the \emph{bookshelf} views as \emph{table} category, and the \emph{bottle} views as \emph{flower pot} category. 
The \emph{bottle} and \emph{flower pot} have close appearances and share the function of holding water. 
Note that the views in the right part (Prediction) are found in the training set and they are hard to distinguish from corresponding input views, even for human beings. 

For 3D shape retrieval, its performance is affected by classification accuracy since 
the misclassified result of a query shape will propagate in the retrieval process, where the model tries to find 
shapes that have the same category as the query. 
Here we visualize the misclassification of several query shapes on the SHREC'17 benchmark, exhibited in the subfigure~\ref{fig:shrec17_retrieval_bad_cases}. 
For instance, the query in the first row is a faucet but our model recognizes it as a lamp. 
The misclassification is somewhat understandable as there are views of \emph{lamp} in the training set with extremely close appearances with \emph{faucet}, seeing the corresponding prediction part. 
Interestingly, in the third row, our model regards two chairs side by side as \emph{sofa}, probably because it learns the common sense that \emph{sofa} is more likely to have consecutive seats than \emph{chair}. 

\subsection{Auxiliary Observations}

\noindent\textbf{Optimizer and Scheduler.}
We examine whether VSFormer is sensitive to the optimizer and lr scheduler by replacing them with common settings, 
such as Adam and CosineAnnealingLR. 
The recognition accuracies on 3 datasets are recorded in Table~\ref{tab:viewformer_adam_coslr}, where the last column 
measures the change relative to the default configuration. We noticed that there was a slight drop in accuracy when 
training with Adam and CosineAnnealingLR. 

\begin{table}[ht]\scriptsize
   \begin{center}
      \caption{The recognition accuracy of VSFormer when optimizing with Adam and CosineAnnealingLR (CosLR). ModelNet40: MN40, ScanObjectNN: SONN.}
      \begin{tabular}{l c c c r l}
      \toprule
      Method & Optim. & Sched. & Dataset & OA (\%) & $\Delta$ (\%)\\
      \midrule
      \multirow{3}{*}{VSFormer} & \multirow{3}{*}{Adam} & \multirow{3}{*}{CosLR} & MN40 & 98.3 & 0.5 $\downarrow$ \\ 
       & & & SONN & 95.9 & 0.0  \\ 
       & & & RGBD & 98.0 & 0.4 $\downarrow$ \\ 
      \bottomrule
      \end{tabular}
      \label{tab:viewformer_adam_coslr}
   \end{center}
\end{table}

\noindent\textbf{Different Methods Using Same Initializer.} 
To be fair, we use the same Initializer for different methods to inspect their recognition accuracies on ModelNet40. 
The chosen methods are strong baselines, RotationNet~\cite{Kanezaki18rotationnet} and View-GCN~\cite{wei20viewgcn}. 
The results in Table~\ref{tab:method_with_different_inits} show VSFormer can achieve higher-level performance no matter 
which initializer is used, 
exceeding View-GCN(AlexNet) and View-GCN(ResNet) by 1.6\% and 1.5\%, respectively. 
The results also indicate the proposed approach is better at grasping multi-view information for recognition since 
the initialized view features are identical. 
\begin{table}[ht]
   \begin{center}
      \caption{Comparison of different multi-view methods with same Initializer.}  %
      \begin{tabular}{l c c}
      \toprule
      Method & Initializer & Inst. Acc. (\%)\\
      \midrule
      RotationNet & \multirow{3}{*}{AlexNet} & 96.4 \\
      View-GCN &  & {97.2} \\
      VSFormer & & \textbf{98.8} \\\hdashline
      RotationNet & \multirow{3}{*}{ResNet50} & 96.9 \\
      View-GCN &  & {97.3} \\
      VSFormer & & \textbf{98.8} \\
      \bottomrule
      \end{tabular}
      \label{tab:method_with_different_inits}
   \end{center}
\end{table}

\noindent\textbf{Shallow Convolutions in Initializer.} 
We investigate the performances of VSFormer when deploying shallow convolution operations as Initializer, \textit{e.g.}, 1- and 2-layer
convolution. Table~\ref{tab:init_shallow_convs} explains their specific configurations. Due to 
the increased number of strides, 2-layer convolution has much lower parameters than 1-layer operation. However, VSFormer with
shallow convolution initializations does not lead to decent 3D shape recognition. 
The best instance
accuracy is 93.7\%, much lower than 98.8\% given by VSFormer with lightweight CNN (AlexNet) Initializer,
suggesting lightweight CNNs are reasonable choices for the Init module.
\begin{table}[ht]
   \begin{center}
   \caption{The configurations of shallow convolutions in Initializer.}  %
      \begin{tabular}{l | r  r}
      \hline
      Conv(s) & \multicolumn{1}{r|}{1-layer} & 2-layer\\
      \hline
      View Size & \multicolumn{2}{c}{$224\times224\times3$} \\
      \hline
       \multirow{4}{*}{1st Conv} & \multicolumn{2}{c}{Conv2d(in=3,out=64,k=7,s=2,p=3)} \\ 
       & \multicolumn{2}{c}{BathNorm2d(num=64)} \\
       & \multicolumn{2}{c}{ReLU(inplace=True)} \\   
       & \multicolumn{2}{c}{MaxPool2d(k=3,s=2,p=1)} \\
      \hline
       \multirow{3}{*}{2nd Conv} & \multicolumn{1}{r|}{\multirow{3}{*}{None}} & Conv2d(in=64,out=32,k=3,s=2,p=1) \\ 
       & \multicolumn{1}{r|}{} & BatchNorm2d(num=32) \\
       & \multicolumn{1}{r|}{} & ReLU(inplace=True) \\    
      \hline
      \#Params (M) & \multicolumn{1}{r|}{{102.8}} & \textbf{12.9} \\
      \hline
      Class Acc. (\%) & \multicolumn{1}{r|}{\textbf{90.1}} & {88.9} \\
      \hline
      Inst. Acc. (\%) & \multicolumn{1}{r|}{{92.5}} & \textbf{93.7} \\
      \hline
      \end{tabular}
   \label{tab:init_shallow_convs}
   \end{center}
\end{table}

\subsection{Limitations}
It is worth noting that VSFormer has some limitations.
First, the Transition module may be a weak point. In many related works, this module is 
designed as a pooling operation or some variant. We employ a concatenation of max 
and mean pooling to summarize the higher-order view correlations into a descriptor, which will inevitably lose 
a part of well-learned correlations. 
Second, it may not be necessary to adopt the same architecture to execute recognition and retrieval tasks. 
3D shape retrieval is more complex as it requires finding highly relevant shapes and 
generating a rank list for them. 
The recognition and retrieval models can share an encoder but vary in the decoder. 
For challenging scenarios, the decoder in the retrieval model can directly operate on the grasped higher-order view 
representations instead of the compressed descriptor. 

\section{Conclusion}
This paper presents VSFormer, a succinct and effective multi-view 3D shape analysis method. 
We organize the different views of a 3D shape into a permutation-invariant set and devise a lightweight attention model 
to capture the correlations of all view pairs. A theoretical analysis is provided to bridge the view set and 
attention mechanism. 
VSFormer shows outstanding performances across different datasets and sets new records for recognition and retrieval tasks. 

In the future, we plan to investigate new paradigms of aggregating the well-learned multi-view correlations without 
losing useful information in the transition. Besides, we are interested in exploring more sophisticated designs 
for the retrieval task and evaluating on diverse benchmarks. As this paper suggested, pairwise similarities are not sufficient to capture
the intrinsic structure of the data manifold~\cite{donoser13diffusion}. 
It is also challenging but worthwhile to extend 3D shape analysis to scene-level tasks, such as 
multi-view 3D semantic segmentation and object detection. 

\section*{Acknowledgments}
Dr. Deying Li is supported in part by the National Natural Science Foundation of China Grant No. 12071478. Dr. Yongcai Wang is supported in part by the National Natural Science Foundation of China Grant No. 61972404, Public Computing Cloud, Renmin University of China, and the Blockchain Lab, School of Information, Renmin University of China.

\bibliographystyle{IEEEtran}
\bibliography{egbib}

\begin{IEEEbiography}[{\includegraphics[width=1in,height=1.25in,clip,keepaspectratio]{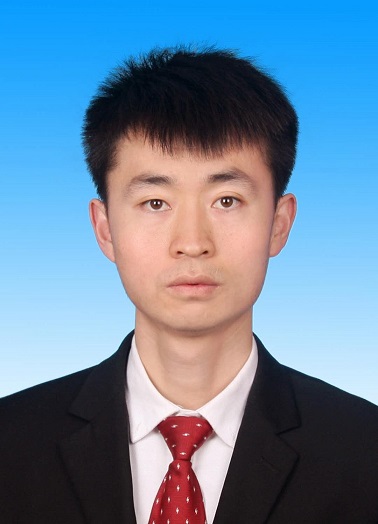}}]{Hongyu Sun}
   obtained his bachelor degree in Computer Science and Technology at School of Computing from Inner Mongolia University, in 2017. He 
   received his master degree in Computer Application Technology at School of Information from Renmin University of China, in 2020. He is currently working toward the Ph.D. degree in the School of Information, Renmin University of China. His research interests include 3D shape analysis and 3D point cloud understanding.
\end{IEEEbiography}
\begin{IEEEbiography}[{\includegraphics[width=1in,height=1.25in,clip,keepaspectratio]{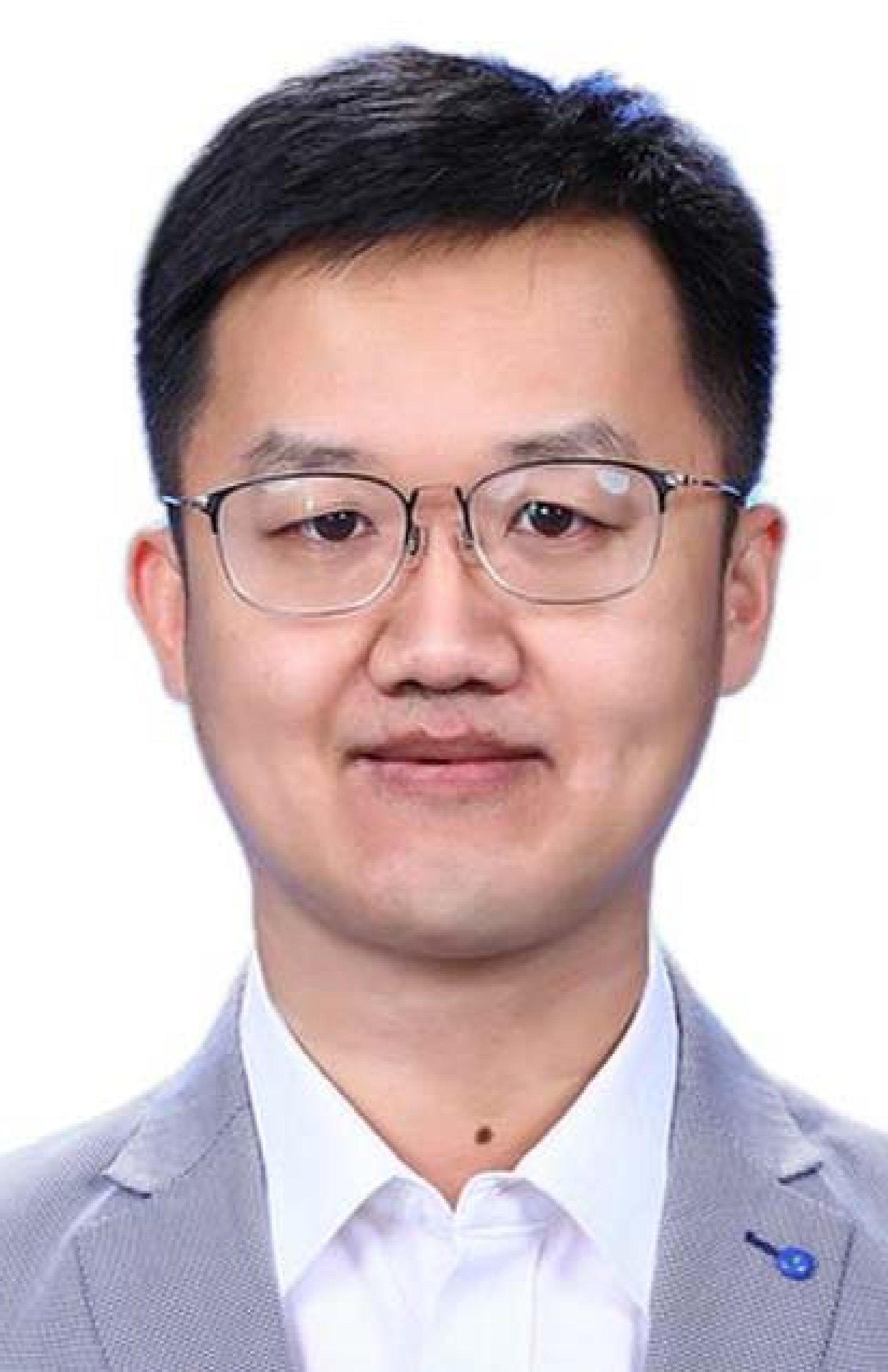}}]{Yongcai Wang}
   received his BS and PhD degrees from Department of Automation Sciences and Engineering, Tsinghua University in 2001 and 2006, respectively. He worked as associated researcher at NEC Labs, China from 2007-2009. He was a research scientist in Institute for Interdisciplinary Information Sciences (IIIS), Tsinghua University from 2009-2015. He was a visiting scholar at Cornell University in 2015. He is currently associate professor at Department of Computer Sciences, Renmin University of China. His research interests include perception and optimization algorithms in intelligent and networked systems. 
\end{IEEEbiography}
\begin{IEEEbiography}[{\includegraphics[width=1in,height=1.25in,clip,keepaspectratio]{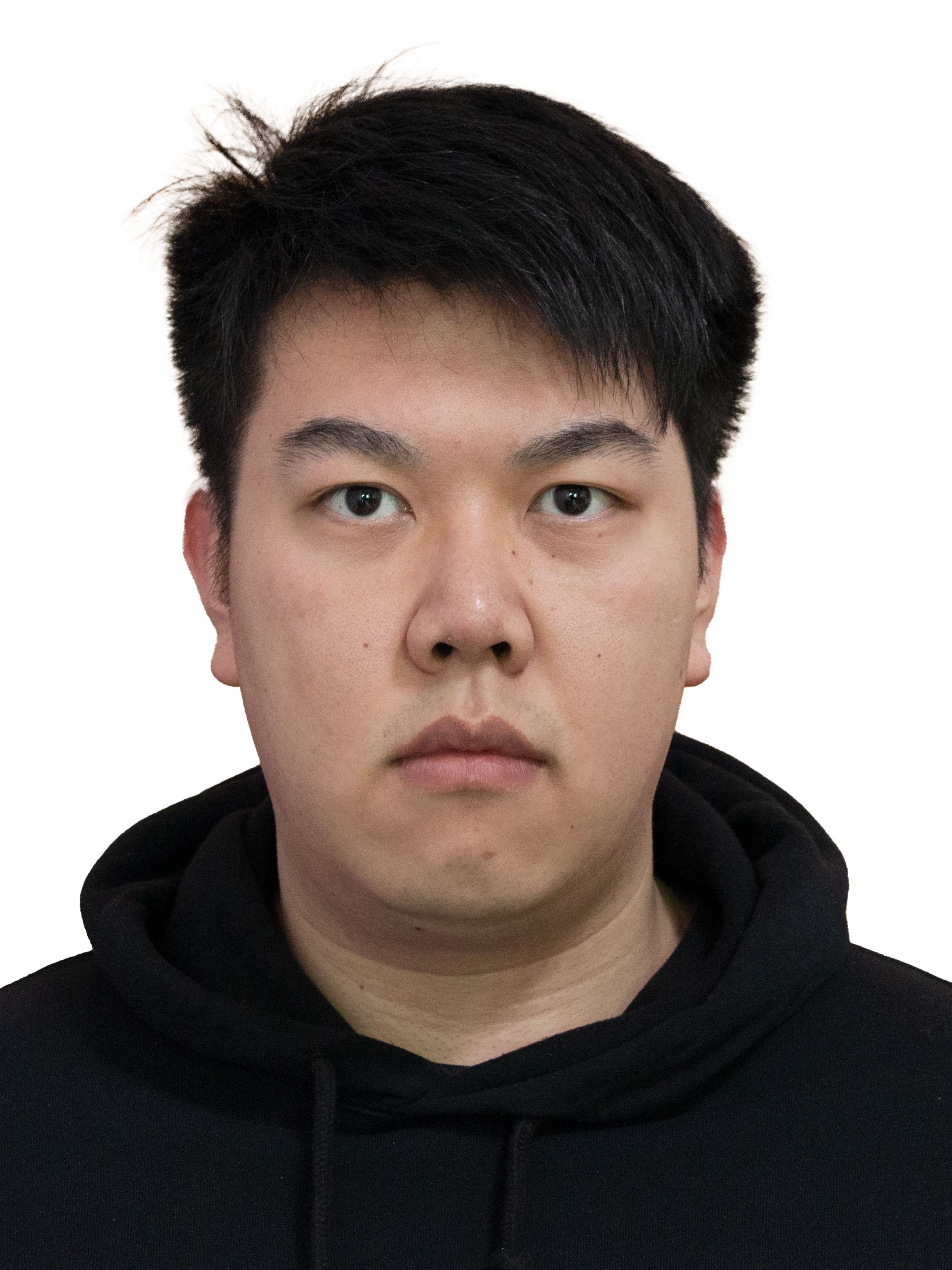}}]{Peng Wang}
   received his bachelor degree in Finance at International School of Business and Finance from Sun Yat-sen University, China, in 2018. He is currently working toward the master degree in the School of Information, Renmin University of China. His research interests include Computer Vision and Multi-Object Tracking.
\end{IEEEbiography}
\begin{IEEEbiography}[{\includegraphics[width=1in,height=1.25in,clip,keepaspectratio]{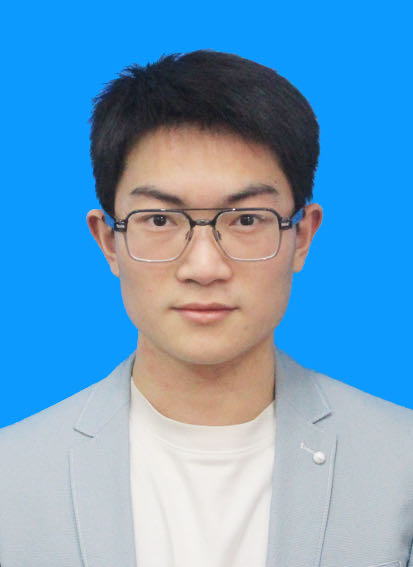}}]{Haoran Deng}
   received his bachelor degree in Computer Science and Technology at School of Remote Sensing Information Engineering from Wuhan University, China, in 2022. He is currently working toward the master degree in the School of Information, Renmin University of China. His research interests include 3D Point Cloud Analysis and Simultaneous Localization and Mapping (SLAM).
\end{IEEEbiography}
\begin{IEEEbiography}[{\includegraphics[width=1in,height=1.25in,clip,keepaspectratio]{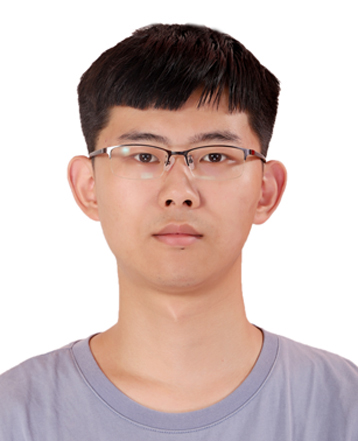}}]{Xudong Cai}
   received his bachelor degree in Computer Science and Technology at School of Computer Science from North China Electric Power University, China, in 2021. He is currently working toward the Ph.D. degree in the School of Information, Renmin University of China. His research interests include Computer Vision and Visual Localization.
\end{IEEEbiography}
\begin{IEEEbiography}[{\includegraphics[width=1in,height=1.25in,clip,keepaspectratio]{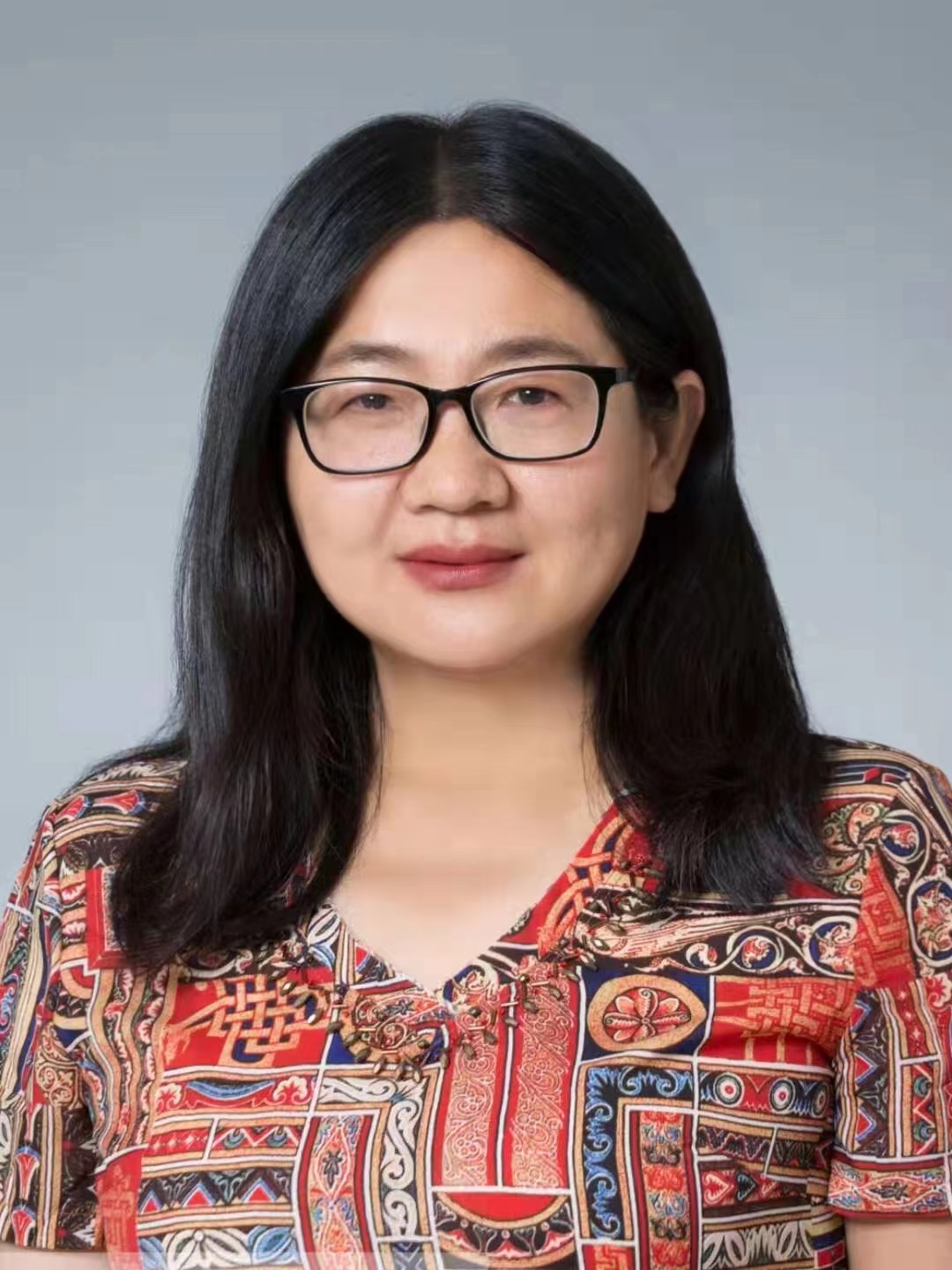}}]{Deying Li}
   received the MS degree in Mathematics from Huazhong Normal University (1988)
   and PhD degree in Computer Science from City University of Hong Kong (2004). She is 
   currently a Professor in Department of Computer Science, Renmin University of China. Her
   research interests include wireless networks, mobile computing, social network and algorithm design and analysis. 
\end{IEEEbiography}

\vfill

\clearpage
\section*{Supplementary Material}
\subsection{View Set Attention Model}

Theorem \ref{theorem:theo1} in Section~\ref{subsubsec:view_set_att_theory}.  
\begin{IEEEproof}
   For a view set $\mathcal{V} = \{v_1, \dots, v_M\}$, let $\mathcal{Z} = \{\textbf{z}_1, \dots, \textbf{z}_M\}$ denote the initialized representations of $\mathcal{V}$. 
   We can derive the Cartesian product of $\mathcal{Z}$, denoted by $\mathcal{P} = \{(\textbf{z}_i,\textbf{z}_j)\ |\ i,j \in 1,\dots,M\}$. 
   Let $p_{i,j} = (\textbf{z}_i,\textbf{z}_j)$, so $\mathcal{P} = \{p_{i,j}\ |\ i,j \in 1,\dots,M\}$. 
   Recall the standard attention model~\cite{vaswani17transformer} that receives the input $\mathcal{I} = \{\textbf{e}_1, \dots, \textbf{e}_N\}$, where the correlation matrix of $\mathcal{I}$ is $\mathcal{A} = \{a_{i,j}\ |\ i,j \in 1,\dots,N\}$ and $a_{i,j}$ represents the attention score that $\textbf{e}_i$ attains from $\textbf{e}_j$. 
   In the attention mechanism, $\mathcal{A}$ can be further decomposed into $\textrm{Norm}(QK^{\textrm{T}}/{\tau})$,
   where $Q = IW_Q$, $K = IW_K$, Norm represents a normalized function ($e.g.$, \verb|softmax|) and $\tau$ is a temperature coefficient. Both $W_Q$ and $W_K$ are learnable parameters in the model.
   Note $\mathcal{P}$ and $\mathcal{A}$ have the same mathematical expression, so we can formulate the Cartesian product $\mathcal{P}$ and model the relations of all pairs in $\mathcal{P}$ by making $N = M$ and $\mathcal{I} = \mathcal{Z}$. 
\end{IEEEproof}

\subsection{Multiple Views of a Retrieved Shape}
\begin{figure}[ht]
	\centering
   \begin{subfigure}{0.2\linewidth}
      \includegraphics[width=\linewidth]{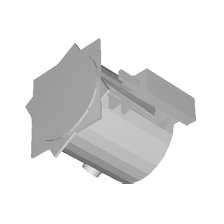}
      \caption{}
   \end{subfigure}%
   \begin{subfigure}{0.2\linewidth}
      \includegraphics[width=\linewidth]{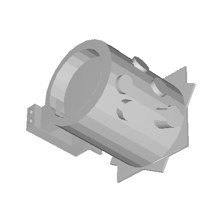}
      \caption{}
   \end{subfigure}%
   \begin{subfigure}{0.2\linewidth}
      \includegraphics[width=\linewidth]{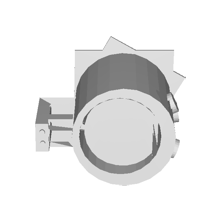}
      \caption{}
   \end{subfigure}%
   \begin{subfigure}{0.2\linewidth}
      \includegraphics[width=\linewidth]{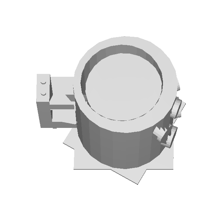}
      \caption{}
   \end{subfigure}%
   \begin{subfigure}{0.2\linewidth}
      \includegraphics[width=\linewidth]{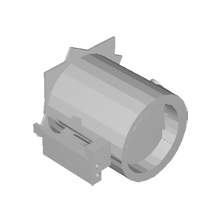}
      \caption{}
   \end{subfigure}\\
   \begin{subfigure}{0.2\linewidth}
      \includegraphics[width=\linewidth]{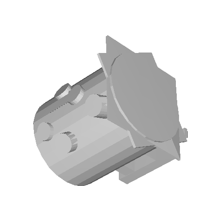}
      \caption{}
   \end{subfigure}%
   \begin{subfigure}{0.2\linewidth}
      \includegraphics[width=\linewidth]{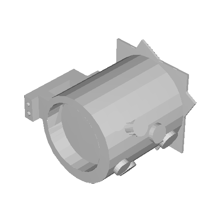}
      \caption{}
   \end{subfigure}%
   \begin{subfigure}{0.2\linewidth}
      \includegraphics[width=\linewidth]{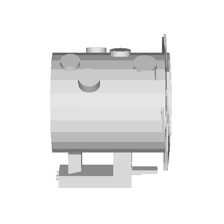}
      \caption{}
   \end{subfigure}%
   \begin{subfigure}{0.2\linewidth}
      \includegraphics[width=\linewidth]{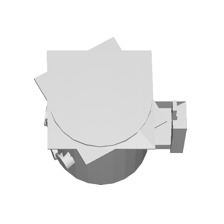}
      \caption{}
   \end{subfigure}%
   \begin{subfigure}{0.2\linewidth}
      \includegraphics[width=\linewidth]{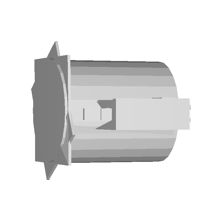}
      \caption{}
   \end{subfigure}%
	\caption{Different views of a retrieved shape.}
   \label{fig:views_of_a_cup}
\end{figure}

In Figure \textcolor{red}{6} of the main paper, the retrieved shape in the 5th column of the 3rd row may be confusing since 
one may not be able to determine whether it belongs to the same class as the query. 
To this end, we pinpoint the shape in the dataset and find more views of it, shown in Figure~\ref{fig:views_of_a_cup}. 
After observing these views, we can infer this shape is a cup, so it is of the same class as the query. 

\subsection{3D Shape Retrieval}
We explore the performances of 3D shape retrieval on the perturbed version of SHREC'17 and
the results are exhibited in Table \ref{tab:ret_shrec17_perturbed}. 
VSFormer leads the strongest baseline View-GCN++ in 9 out of 10 metrics on the challenging dataset. 
The advantages are notable in macro-version metrics, achieving 2.9\% absolute improvements on average. 
NDCG is a widely used metric in information retrieval, which penalizes poor rankings of a retrieval list. 
We lag behind View-GCN++ by 2.4\% in micro NDCG, which is probably because the simple transition operation 
compresses the informative correlations learned by the encoder, so the model is confused by similar shapes 
and difficult to decide their rankings. 

\begin{table}[t]
   \begin{center}
   \caption{Comparison of 3D shape retrieval on the perturbed version of ShapeNet Core55.}
   \begin{tabular}{l | c c c c c }
      \toprule
      Method & P@N & R@N & F1@N & mAP & NDCG \\
      \midrule
      & \multicolumn{5}{c}{\emph{micro}} \\
      GIFT~\cite{bai16gift} & 67.8 & 66.7 & 66.1 & 60.7 & 73.5 \\
      Improved GIFT~\cite{bai17gift} & 66.0 & 65.0 & 64.3 & 56.7 & 70.1 \\
      REVGG & 70.5 & 76.9 & 71.9 & 69.6 & 78.3 \\
      CM-VGG5-6DB & 41.2 & 70.6 & 47.2 & 52.4 & 64.2 \\
      MVCNN~\cite{su15mvcnn} & 63.2 & 61.3 & 61.2 & 53.5 & 65.3 \\
      RotationNet~\cite{Kanezaki18rotationnet} & 65.5 & 65.2 & 63.6 & 60.6 & 70.2 \\
      View-GCN~\cite{wei20viewgcn} & 75.4 & 75.0 & 74.6 & 72.2 & 78.9 \\
      View-GCN++~\cite{wei23viewgcn2} & 76.1 & 75.5 & 75.2 & 72.6 & \textbf{79.8} \\\hdashline
      \textbf{VSFormer} & \textbf{76.8} & \textbf{78.1} & \textbf{76.8} & \textbf{73.7} & 77.4 \\
      \midrule
      & \multicolumn{5}{c}{\emph{macro}} \\
      GIFT~\cite{bai16gift} & 41.4 & 49.6 & 42.3 & 41.2 & 51.8 \\
      Improved GIFT~\cite{bai17gift} & 44.3 & 50.8 & 43.7 & 40.6 & 51.3 \\
      REVGG & 42.4 & 56.3 & 43.4 & 41.8 & 47.9 \\
      CM-VGG5-6DB & 12.0 & 65.9 & 16.4 & 32.9 & 39.5 \\
      MVCNN~\cite{su15mvcnn} & 40.5 & 48.4 & 41.5 & 36.7 & 45.9 \\
      RotationNet~\cite{Kanezaki18rotationnet} & 37.2 & 39.3 & 33.3 & 32.7 & 40.7 \\
      View-GCN~\cite{wei20viewgcn} & 52.4 & 57.3 & 52.6 & 49.6 & 54.8 \\
      View-GCN++~\cite{wei23viewgcn2} & 53.8 & 57.7 & 53.5 & 50.1 & 55.5 \\\hdashline
   \textbf{VSFormer} & \textbf{56.0} & \textbf{59.3} & \textbf{55.7} & \textbf{53.0} & \textbf{61.1} \\
      \bottomrule
   \end{tabular}
   \label{tab:ret_shrec17_perturbed}
   \end{center}
\end{table}

\end{document}